\documentclass{article} 
\usepackage{iclr2024_conference,times}
\iclrfinalcopy

\usepackage{amsmath,amsfonts,bm}

\def\eqref#1{equation~\ref{#1}}

\def\1{\bm{1}}

\DeclareMathAlphabet{\mathsfit}{\encodingdefault}{\sfdefault}{m}{sl}
\SetMathAlphabet{\mathsfit}{bold}{\encodingdefault}{\sfdefault}{bx}{n}

\usepackage{hyperref}
\usepackage{url}
\usepackage{float}
\usepackage{caption}
\usepackage{subcaption}
\usepackage{graphicx}
\usepackage{booktabs}
\usepackage{multirow}

\title{Evaluation of post-hoc interpretability methods in time-series classification \thanks{This article was published in Nature Machine Intelligence and is available online at \url{https://doi.org/10.1038/s42256-023-00620-w}}}

\author{Hugues Turbé\thanks{Email: hugues.turbe@unige.ch; mpegim@nus.edu.sg} , Mina Bjelogrlic, Christian Lovis  \\
Division of Medical Information Sciences, University Hospitals of Geneva, Geneva, Switzerland\\
Department of Radiology and Medical Informatics, University of Geneva, Geneva, Switzerland\\
\And
Gianmarco Mengaldo \footnotemark[2] \\
Department of Mechanical Engineering, College of Design and Engineering, \\
National University of Singapore, Singapore, Singapore \\
Honorary Research Fellow, Department of Aeronautics, Imperial College London, London, UK \\
}

\begin{document}

\maketitle

\begin{abstract}
Post-hoc interpretability methods are critical tools to explain neural-network results. Several post-hoc methods have emerged in recent years, but when applied to a given task, they produce different results, raising the question of which method is the most suitable to provide correct post-hoc interpretability. To understand the performance of each method, quantitative evaluation of interpretability methods is essential. However, currently available frameworks have several drawbacks which hinders the adoption of post-hoc interpretability methods, especially in high-risk sectors. In this work, we propose a framework with quantitative metrics to assess the performance of existing post-hoc interpretability methods in particular in time series classification. We show that several drawbacks identified in the literature are addressed, namely dependence on human judgement, retraining, and shift in the data distribution when occluding samples. We additionally design a synthetic dataset with known discriminative features and tunable complexity. The proposed methodology and quantitative metrics can be used to understand the reliability of interpretability methods results obtained in practical applications. In turn, they can be embedded within operational workflows in critical fields that require accurate interpretability results for e.g., regulatory policies.
\end{abstract}

\section{Introduction}
\label{sec:introduction}
\begin{figure}[H]
     \centering
     \begin{subfigure}[b]{0.49\textwidth}
         \centering
         \includegraphics[width=\textwidth]{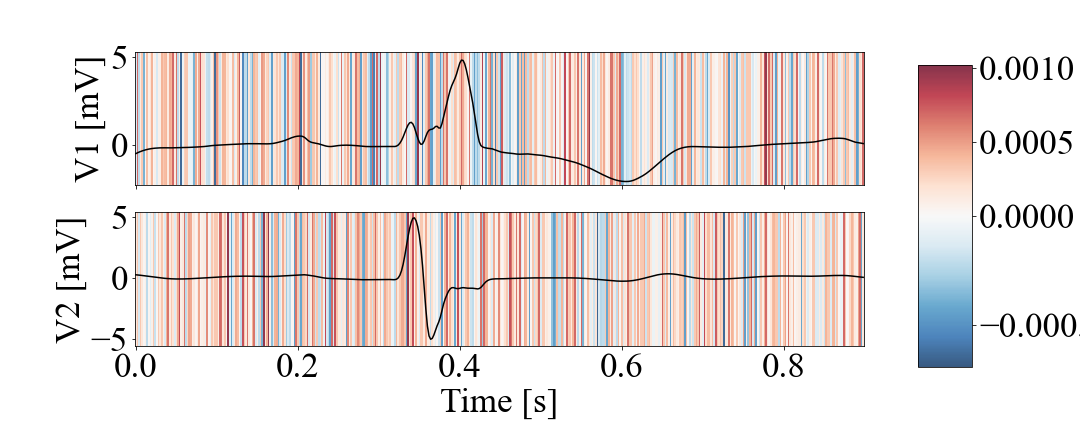}
         \vspace{-0.5cm}
         \caption{KernelShap}
     \end{subfigure}
     \hfill
     \begin{subfigure}[b]{0.49\textwidth}
         \centering
         \includegraphics[width=\textwidth]{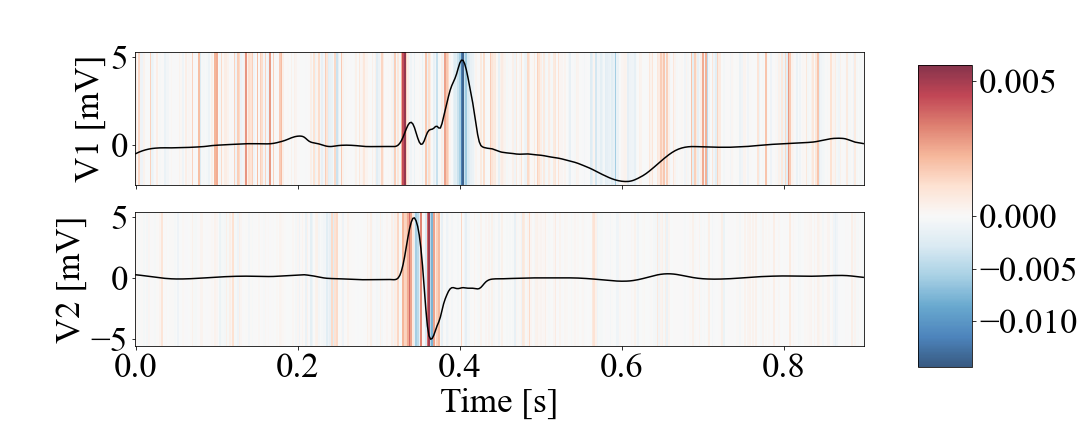}
         \vspace{-0.5cm}
         \caption{Deeplift}
     \end{subfigure}
     \hfill
     \begin{subfigure}[b]{0.49\textwidth}
         \centering
         \includegraphics[width=\textwidth]{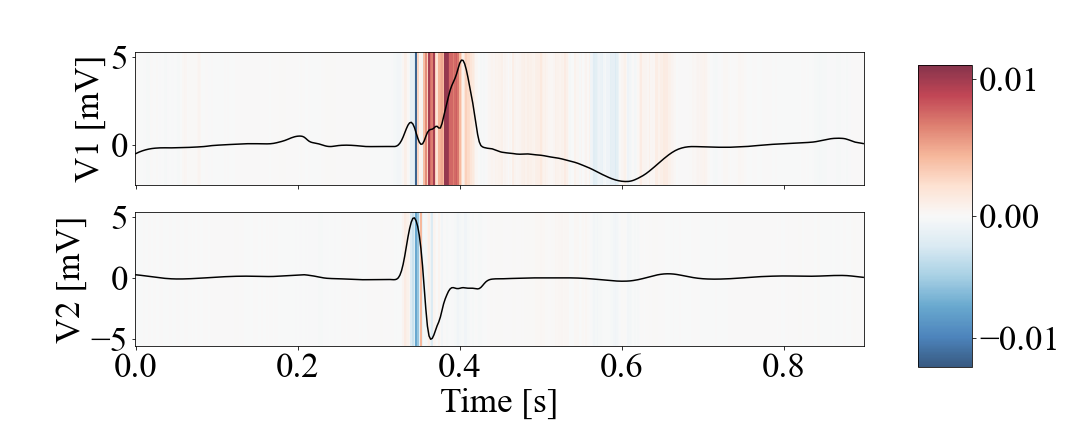}
         \vspace{-0.5cm}
         \caption{DeepliftShap}
     \end{subfigure}
     \hfill
     \begin{subfigure}[b]{0.49\textwidth}
         \centering
         \includegraphics[width=\textwidth]{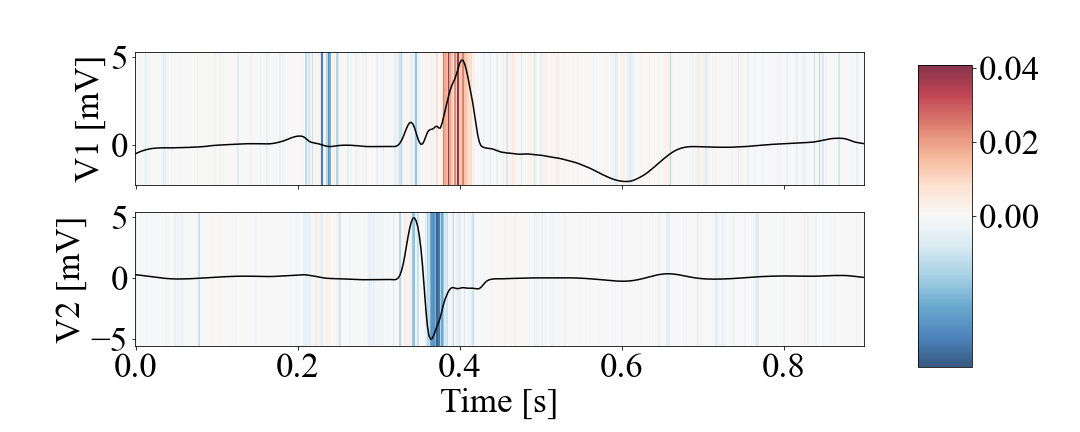}
         \vspace{-0.5cm}
         \caption{Shapley Value}
     \end{subfigure}
        \caption{\textbf{Relevance produced by four post-hoc interpretability methods, obtained on a time-series classification task, where a Transformer neural network needs to identify the pathology of a patient from ECG data. Depicted in black are two signals, V1, and V2, while the contour map represents the relevance produced by the interpretability method.} Red indicates positive relevance, while blue indicate negative relevance. The former marks portions of the time series that were deemed important by the interpretability method for the neural-network prediction. The latter marks portions of the time series that were going against the prediction.}
        \label{fig:differences_interp}
\end{figure}

Time series, sequences of indexed data that follow a specific time order, are ubiquitous. 
They can describe physical systems~\cite{weyn_improving_2020}, such as the state of the atmosphere and its evolution, social and economic systems~\cite{yang_big_2020}, such as the financial market, and biological systems~\cite{rajkomar2018scalable}, such as the heart and the brain via electrocardiogram (ECG) and electroencephalogram (EEG) signals, respectively. 
Availability of this type of data is increasing, and so is the need for automated analysis tools capable of extracting interpretable and actionable knowledge from them. To this end, while established, and more interpretable time-series approaches remain competitive for many tasks~\cite{dau2019ucr,manibardo2021deep, ye2009time}, artificial intelligence (AI) technologies, and neural networks in particular, are opening the path towards highly-accurate predictive tools for an increasing number of time-series regression~\cite{hewamalage2021recurrent,lim2021temporal, tang2021probabilistic} and classification~\cite{fawaz2019deep,hong2022practical} learning tasks. 
Yet, the adoption of AI technologies as black-box tools is problematic in several applied contexts. 
To address this issue, numerous interpretability methods have been proposed in the literature, especially in the context of neural networks. 
These different methods usually produce tangibly different results, preventing practitioners to fully unlock the interpretability of the results increasingly needed. 
Figure~\ref{fig:differences_interp} shows four different post-hoc interpretability methods applied to time-series classification, where the neural network is tasked with identifying the pathology associated to a patient's ECG.
The four interpretability methods produce remarkably different results for the same model. 
Hence, the question: \textit{what method produced an interpretability map closer to the one actually adopted by the neural network to make its prediction?} 
Answering this question quantitatively, while addressing the issues found in the existing literature on interpretability methods' evaluation, is the scope of this paper.
Indeed, aside from research purposes, understanding the accuracy of interpretability methods is de facto mandatory in critical sectors (e.g., healthcare) for legal and ethical reasons~\cite{ai_act}. 
Failing to understand the performance of interpretability methods may prevent their adoption, that in turn can lead practitioners to avoid using neural-network tools altogether, in favor of more white-box and interpretable tools. 

Different definitions of what it means for a neural-network model to be interpretable have been formulated. 
Most of these definitions can be summarised under two categories: \textit{transparency} and \textit{post-hoc interpretability}~\cite{lipton_mythos_2017}.
Transparency refers to how a model and its individual constituents work. 
Post-hoc interpretability refers to how a trained model makes predictions and uses the input features it is given. 
In this work, we consider post-hoc interpretability applied to time series classification, because it is seen as a key to meet recent regulatory requirements~\cite{ai_act} and translate current research efforts into real-world applications, especially in high risk areas, such as healthcare~\cite{shad_designing_2021}. 
Post-hoc interpretability methods assign a relevance to each feature of a sample reflecting its importance to the model for the classification task being performed.
The ability to express the specific features used by a neural network to classify a given sample can help humans assess the reliability of the classification produced and allows comparing the model's predictions with existing knowledge. 
It also provides a way to understand possible model's biases which could lead to the failure of the model in a real-world setting.  

A range of methods to provide post-hoc interpretability have been developed in the past few years to interpret classification results. 
These are mainly focused on natural language processing (NLP) and image classification tasks. 
More recently, with the growing interest for neural-network interpretability, leading actors in the machine learning community built a range of post-hoc interpretability methods. 
As part of this effort, Facebook recently released the Captum library to group a large number of interpretability methods under a single developmental framework~\cite{kokhlikyan2020captum}.  
While these initiatives allow researchers to use the different interpretability methods more easily, they do not provide a systematic and comprehensive evaluation of those methods on data with different characteristics and across neural-network architectures. 
A systematic methodology that provides the accurate evaluation of these methods is of paramount importance to allow their wider adoption, and measure how trustable the results they provide are.

The evaluation of interpretability methods was initially based on a heuristic approach, where the relevance attributed to the different features was compared to the expectation of an observer for common image classification tasks~\cite{montavon_explaining_2017}, or of a domain expert for more complex tasks~\cite{NIPS2017_8a20a862, neves2021interpretable}. 
However, these works had a common pitfall: they assumed the representation of a task learned by a neural network should use the same features as a human expert. 
The community later moved towards the idea that the evaluation should be independent of human judgement~\cite{jacovi2020towards}. 
This paradigm shift was supported by the evidence that certain saliency methods, while looking attractive to human experts, produced results independent of the model they aimed to explain, thereby failing the interpretability task~\cite{adebayo2018sanity}.
More recent evaluations were performed by occluding (also referred to as corrupting) the  most relevant features identified and comparing the drop in score observed between model's predictions on the initial and modified samples~\cite{samek2016evaluating}. 
This evaluation method was later questioned, as corrupting the images changes the distribution of the sample's values and therefore the observed drop in score might be caused by this shift in distribution rather than actual information being removed~\cite{hooker_benchmark_2019}. 
To address this issue, an approach named ROAR was proposed~\cite{hooker_benchmark_2019}, where important pixels are removed both in the train and test set. The model is then retrained on the corrupted (i.e., occluded) samples, with the drop in score being retained on this newly trained model. 
This method has the benefit of maintaining a similar distribution across the train and evaluation set with the modified samples. 
Yet, we argue that it does not necessarily explain which features the initial network used to make its prediction as the similarity between neural network models is only maintained if the models are trained on datasets sampled from the same distribution \cite{hacohen2020let}. 
In their case, the distribution is changed as the model is retrained on a corrupted dataset and therefore the post-hoc interpretability of the retrained model is not constrained to being similar to the one of the initial model. 
The post-hoc interpretability rather highlights the properties of the dataset in regards to its target, such as the redundancy of the information present in the features that are indicative of a given class -- limitation that was acknowledged by the authors~\cite{samek2016evaluating}.

Neural-network interpretability for time-series data was only recently explored. 
Initial efforts applied some of the interpretability methods introduced for NLP and image classification on univariate time series, and evaluated the drop in score obtained by corrupting the most relevant parts (also referred to as time steps) of the signal~\cite{schlegel2019towards}.  
An evaluation of some interpretability methods was recently proposed~\cite{ismail_benchmarking_2020}, with a dataset designed to address the issue of retaining equal distribution between the initial and the occluded dataset.
However, this work may have two crucial drawbacks: the proposed dataset contains static discriminative properties (e.g., the mean of the sample), and it is not independent of human judgement. 
The first issue can lead the model to learn from static properties. 
Hence the dataset might not reflect the complexity of real-world time-series classification tasks, where time dependencies usually play the discriminative role. 
The second issue is related to the assumption that the model uses: \begin{itemize}
    \item all the discriminative information synthetically provided (that consist of a static shift applied to a portion of the time series),
    \item no information outside of it.
\end{itemize} 
We argue that this assumption does not necessarily hold as the model might require just a subset of the discriminative information provided and might use information from outside the discriminative portion.

In this work, we propose an approach for the model-agnostic evaluation of interpretability methods for time series classification, that addresses the various issues just highlighted. 
The approach consists of two new metrics, namely $AUC\tilde{S}_{\textrm{top}}$, and $F1\tilde{S}$. 
The first aim to measure how the top relevance indeed captures the most important time steps for the neural network. 
The second is a harmonic mean reflecting both the capability of the different interpretability methods to capture the most important time steps  as well as the least important ones. 
 These two metrics evaluate how interpretability methods order time steps  according to their importance referred to as \textit{relevance identification}. In this paper, we also aim to qualitatively evaluate the capacity of the different interpretability methods to reflect the importance of each time step relative to the others. The latter evaluation is referred to as \textit{relevance attribution}. 
We note that a key aspect of the work is the training of the models with a random level of perturbation for each batch, in a similar fashion as widely used data augmentation methods~\cite{Liu_2018_CVPR}. 
This perturbation is later used to corrupt the signal when evaluating interpretability methods such that the distribution is maintained across the training dataset and perturbed dataset used for the evaluation. 
This addresses one of the main concerns found in the literature, that is the shift in distribution when occluding samples in the evaluation set, and does not need retraining as the ROAR approach.

The interpretability methods we considered are six, namely: i) DeepLift~\cite{deeplift}, ii) GradShap~\cite{NIPS2017_8a20a862}, iii) Integrated Gradients~\cite{ig_method}, iv) KernelShap~\cite{NIPS2017_8a20a862}, v) DeepListShap~\cite{NIPS2017_8a20a862}, vi) Shapley Value Sampling (also referred to as Shapley Sampling, or simply Shapley)~\cite{CASTRO20091726}. 
These were chosen to capture a broad range of available interpretability methods, while maintaining the problem computationally tractable for all the models presented. 
These interpretability methods are applied to three neural-network architectures, namely convolutional (CNN), bidirectional long-short term memory (Bi-LSTM), and Transformer neural networks. 
The evaluation of the interpretability methods for time-series classification is carried out on a new synthetic dataset as well as on two datasets adopted in practical applications. 
The overall code framework is part of the InterpretTime library freely available in Github\footnote{\url{https://github.com/hturbe/InterpretTime}}.

In summary, the approach proposed and the new synthetic dataset we just outlined address the following points:
\begin{enumerate}
\item The necessity for a robust and quantifiable approach to evaluate and rank interpretability methods' performance over different neural-network architectures trained for the classification of time series. 
Our approach addresses the issues found in the literature by providing novel quantitative metrics for the evaluation of interpretability methods independent of human judgement~\cite{jacovi2020towards}, using an occluded dataset~\cite{samek2016evaluating}, and without retraining the model~\cite{hooker_benchmark_2019}.
\item The lack of a synthetic dataset with tunable complexity that can be used to assess the performance of interpretability methods, and that is able to reproduce time-series classification tasks of arbitrary complexity. 
We note that our synthetic dataset differs from \cite{ismail_benchmarking_2020}, as the neural network must learn the time dependencies in the data. 
In addition, the dataset encodes a-priori knowledge of the discriminative features, in analogy to the BlockMNIST synthetic dataset~\cite{shah2021input}.   Finally, the classification task is multivariate by design, since the neural network must learn at least two features to predict the correct class. The latter is a desirable property, as real-world datasets are commonly multivariate.
\end{enumerate}

This paper is organized as follows. In section ~\ref{sec:results}, we present the key results. In section~\ref{sec:discussion}, we discuss the results and summarise the main conclusions. 
In methods, we outline the new approach to interpretability evaluation for time series classification, including the novel method used to maintain a constant distribution between the training and evaluation sets (methods~\ref{sec:model_training}), the new metrics (methods~\ref{sec:interpretability}), and the synthetic dataset (methods~\ref{sec:datasets}).

\section{Results}
\label{sec:results}
All metrics presented in this section are built on the relevance that an interpretability method provides along the time series, and denoted with $\mathbf{R}$ (a more detailed explanation for $\mathbf{R}$ is provided in Table~\ref{table:variable_summary}  and in Methods section~\ref{sec:interpretability}).

\begin{table}[htb]
\centering
\begin{tabular}{@{}lll@{}}
\toprule
                      & Symbol              & \multicolumn{1}{c}{Definition} \\ \midrule
General notation      & $\bar{\cdot}$       & Corrupted value                \\
                      & $\tilde{\cdot}$     & Normalized value               \\ \midrule
Variables             & $\mathbf{X}$        & Sample                         \\
                      & $\mathbf{\bar{X}}$  & Corrupted sample               \\
                      & $\mathbf{R}$        & Attributed relevance            \\
                      & $x_{m,t}$           & Feature $m$ at time step $t$ from a sample $\mathbf{X}$ \\
                      & $r_{m,t}$           & Relevance for feature $m$ at time step $t$ from \\
                      & & a relevance matrix $\mathbf{R}$ \\
                      &  $\mathbf{\bar{X}}^{\text{top}}_k$ & Sample with top-$k$ time steps corrupted    \\
                      &  $\mathbf{\bar{X}}^{\text{bottom}}_k$  & Sample with bottom-$k$ time steps corrupted \\
                      & $R^+$               & Set of positive attributed relevance in a given sample     \\
                      & $M$                  & Number of features in $\mathbf{X}$            \\
                      & $N_c$                & Number of labels                     \\
                      & $J$                  & Number of samples available          \\
                      & $T$                  & Number of ordered time steps per feature  \\
                      & $N$                  & Number of total time steps in given sample ($M\times T$) \\
                      & $\bar{N}$            & Number of time steps corrupted in a given sample           \\
                      & $\tilde{N}$          & Fraction of corrupted time steps in a given sample ($ \frac{\bar{N}}{N}$) \\
                      & $k$                   & Percentage of time steps corrupted with respect to   \\
                      &                     & the total number of time steps with positive relevance \\
                       \midrule
Operators             & $S(\cdot)$           & Post-softmax model's output      \\
                      & $\tilde{S}(\cdot)$   & Normalized change in score (eq.~\ref{eq.s_tilde})     \\ 
                      & $\tilde{S}_A(\cdot)$ & Adjusted $\tilde{S}$ (eq.~\ref{eq:mod-S_A})          \\ 
                      & $\mathcal{A}(\cdot)$ & Attribution scheme               \\ \midrule
Metrics               & $AUC\tilde{S}$       & Area under the $\tilde{S}$ curve (eq.~\ref{eq:aucs}) \\
                      & $F1\tilde{S}$        & Modified F1 Score (eq.~\ref{eq:F1-aucse})        \\ 
                      & $TIC$                & Time information content (eq.~\ref{eq:tic})  \\ 
                      & $IR$                 & Information ratio (eq.~\ref{eq.IR})          \\ \bottomrule
\end{tabular}

\caption{Notations and symbols used in this paper.}
\label{table:variable_summary}
\end{table}

\begin{figure}[htb]
    \centering
    \includegraphics[width=0.6\textwidth]{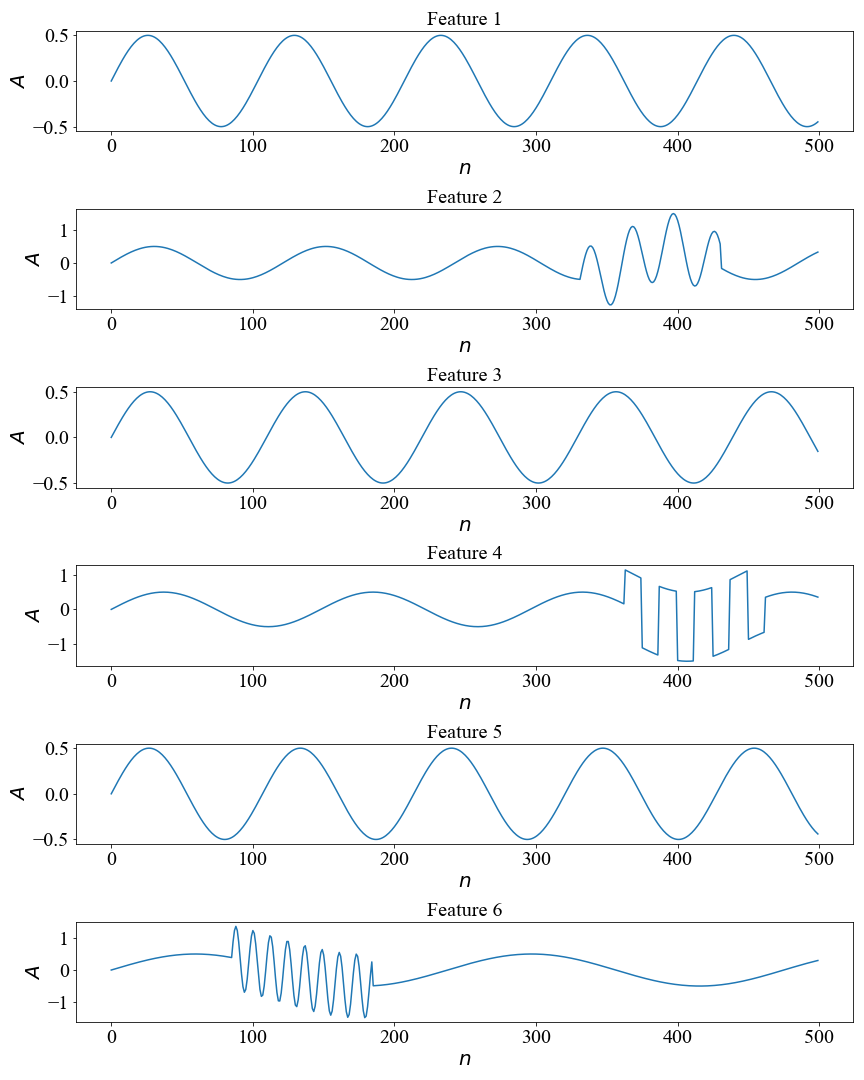}
    \caption{Sample from the synthetic dataset.}
    \label{fig:synthetic_data}
\end{figure}

An example for ECG time series is depicted in figure~\ref{fig:differences_interp}, where the countour map represents the relevance $\mathbf{R}$, while the black line is the actual time series the neural network is using to make the prediction. 
The higher the relevance, the more important the portion of the time series associated with it is for the neural network classification task. The metrics are evaluated on three different datasets, namely synthetic, ECG and FordA, and three different architectures: Bi-LSTM, CNN and Transformer. Models' hyperparameters and the classification metrics for the different models are presented in Supplementary information, section 2 and 3. A sample from the synthetic dataset can be found in figure~\ref{fig:synthetic_data}.
In the following, we focus on evaluating the effectiveness of an interpretability method in ordering time steps according to their importance to explaining the neural network's predictions. 
This crucial aspect of interpretability methods' evaluation is also referred to as \textit{relevance identification}, and it is measured by the two proposed metrics $AUC\tilde{S}_{\textrm{top}}$ and $F1\tilde{S}$ described in Methods section~\ref{sec:interpretability}.

\begin{figure}[htb]
         \centering
         \includegraphics[width=\textwidth]{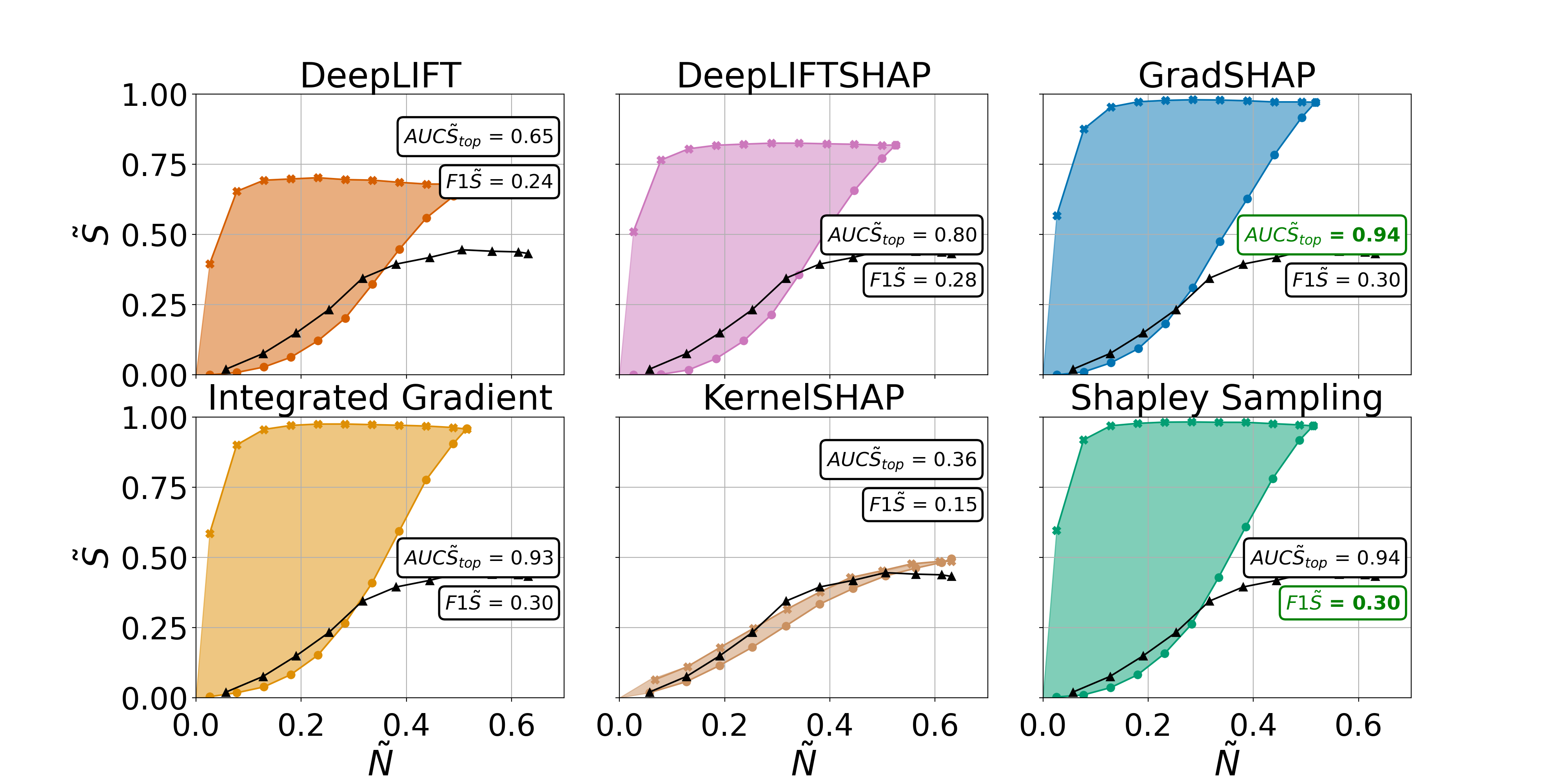 }
        \caption{\textbf{$\tilde{S}$ as a function of the ratio of points removed with respect to the total number of time steps in the sample, $\tilde{N}$.} Each subfigure represents one of the six interpretability methods considered for a transformer trained on the synthetic dataset.}
        \label{fig:syntethic-dataset-identification_transformer}
\end{figure}

The ordering of the time steps obtained using the relevance is used to corrupt the top-$k$ and bottom-$k$ `elements' with positive relevance. Here, with `$k$ elements', we refer to the percentage of time steps in the time series that are corrupted with respect to the total number of time steps with positive relevance. `Top-$k$ elements' refers to a corruption strategy that corrupts time steps starting with higher relevance descending to lower relevance. Similarly, `bottom-$k$ elements', refers to a corruption strategy starting from time steps with low relevance ascending to higher relevance. We note that $k$ is only used for calculating the number of elements to corrupt. However, the evaluation of the interpretability methods is performed with respect to the total number of elements in the sample, denoted by $\tilde{N}$. This was done such that the evaluation of different interpretability methods is independent of the number of time steps assigned with positive relevance, and instead is based on the total number of time steps. Figure~\ref{fig:syntethic-dataset-identification_transformer} shows $\tilde{S}$, the normalized change in score (eq.~\ref{eq.s_tilde} in Methods section~\ref{sec:methods_evaluation}) for a Transformer trained on the newly created synthetic dataset. Results for a Bi-LSTM and CNN architecture are presented in extended data figures~\ref{fig:syntethic-dataset-identification_bilstm} and ~\ref{fig:syntethic-dataset-identification_cnn}.
These $\tilde{S} - \tilde{N}$ curves constitute the basis for computing the $AUC\tilde{S}_{\textrm{top}}$ and $F1\tilde{S}$ metrics. The figure contains all the six interpretability methods considered in this work and a baseline (depicted in black). The latter illustrates $\tilde{S} - \tilde{N}$ for a random assignment of the relevance. Similar figures for the ECG dataset are presented respectively in extended data figures~\ref{fig:ecg-dataset-identification_bilstm},~\ref{fig:ecg-dataset-identification_cnn},~\ref{fig:ecg-dataset-identification_transformer} while results obtained on the FordA dataset are in Supplementary information, section 1.1.
Both, $\tilde{S}$ and $\tilde{N}$ are detailed in methods~\ref{sec:methods_evaluation}. 
As mentioned above, the points $\tilde{N}$ are removed in two ways: from the most important to the least important points identified by the interpretability method (top-$k$ strategy), resulting in the top curve (curve with crosses), and from the least important to the most important points (bottom-$k$ strategy) resulting in the bottom curve (curve with circles).

The $AUC\tilde{S}_{\textrm{top}}$ metric is the area under the top curve: the higher it is the better the interpretability method has understood which points were the most important for the model to assign the correct class. 
The smaller the area under the bottom curve, the better the interpretability method has understood which points were the least important for the model to assign the correct class.
Therefore, a good trade-off between the two shows the interpretability method has identified both the most and the least important points. 
The $F1\tilde{S}$ metric represents the harmonic mean between the capacity to extract the most relevant time steps as well as the least relevant ones. 
A higher score, as for the $AUC\tilde{S}_{\textrm{top}}$ metric, represents a better \textit{relevance identification} performance. 

\begin{figure}[htb]
     \centering
     \begin{subfigure}[b]{0.49\textwidth}
         \centering
         \includegraphics[width=\textwidth]{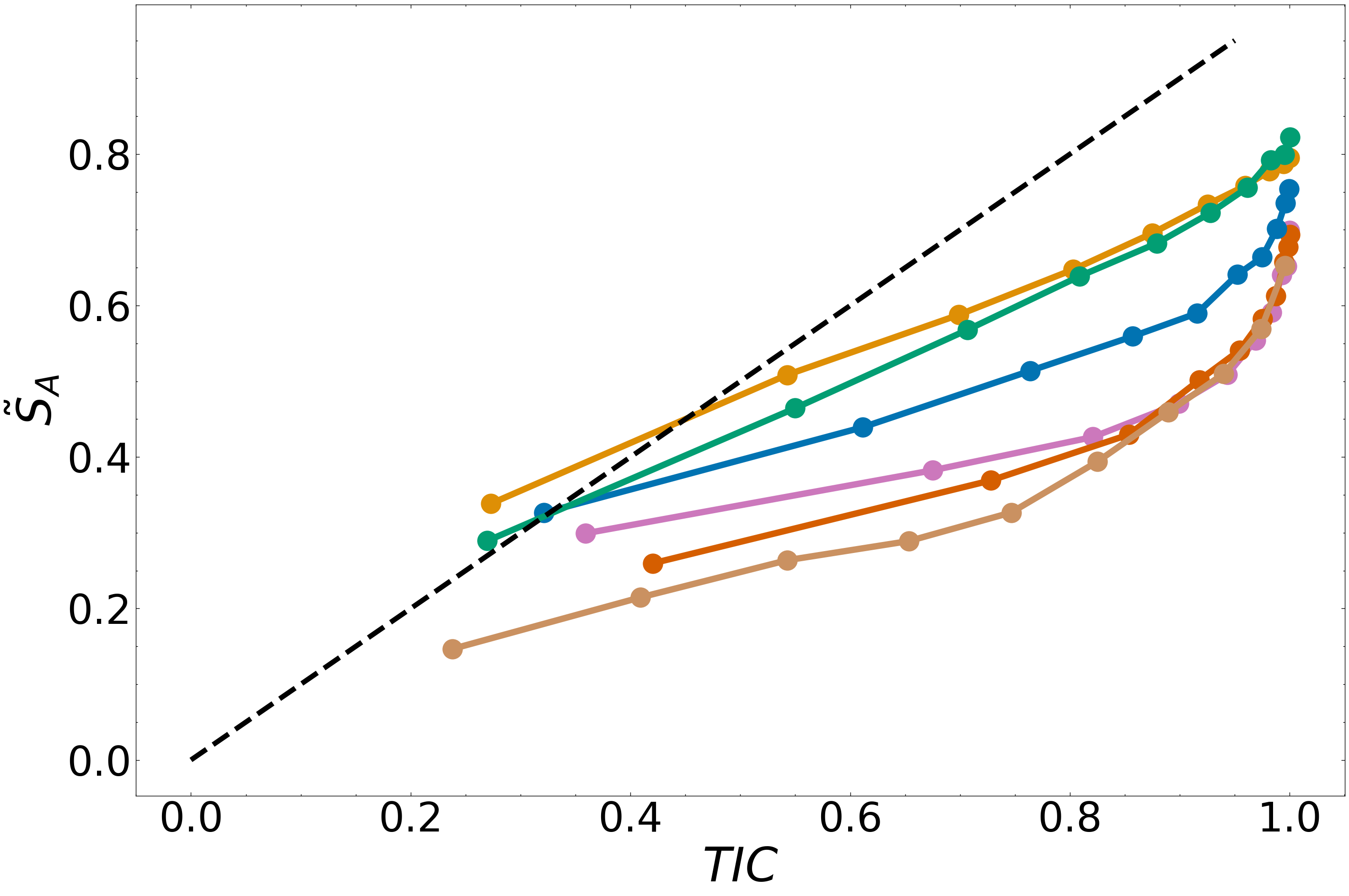}
         \vspace{-0.5cm}
         \caption{Bi-LSTM }
     \end{subfigure}
     \hfill
     \begin{subfigure}[b]{0.49\textwidth}
         \centering
         \includegraphics[width=\textwidth]{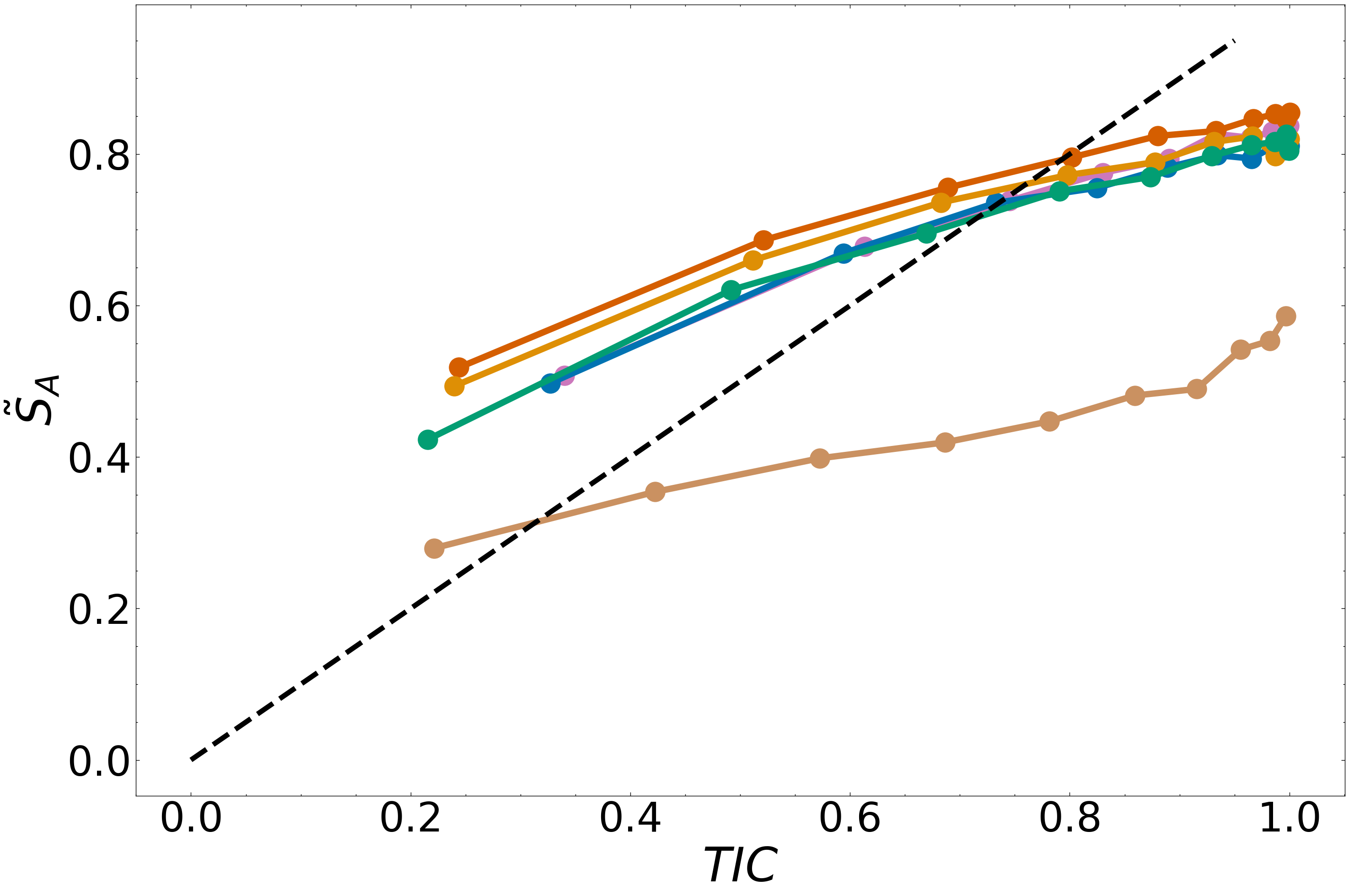}
         \vspace{-0.5cm}
         \caption{CNN}
     \end{subfigure}
     \hfill
     \begin{subfigure}[b]{0.49\textwidth}
         \centering
         \includegraphics[width=\textwidth]{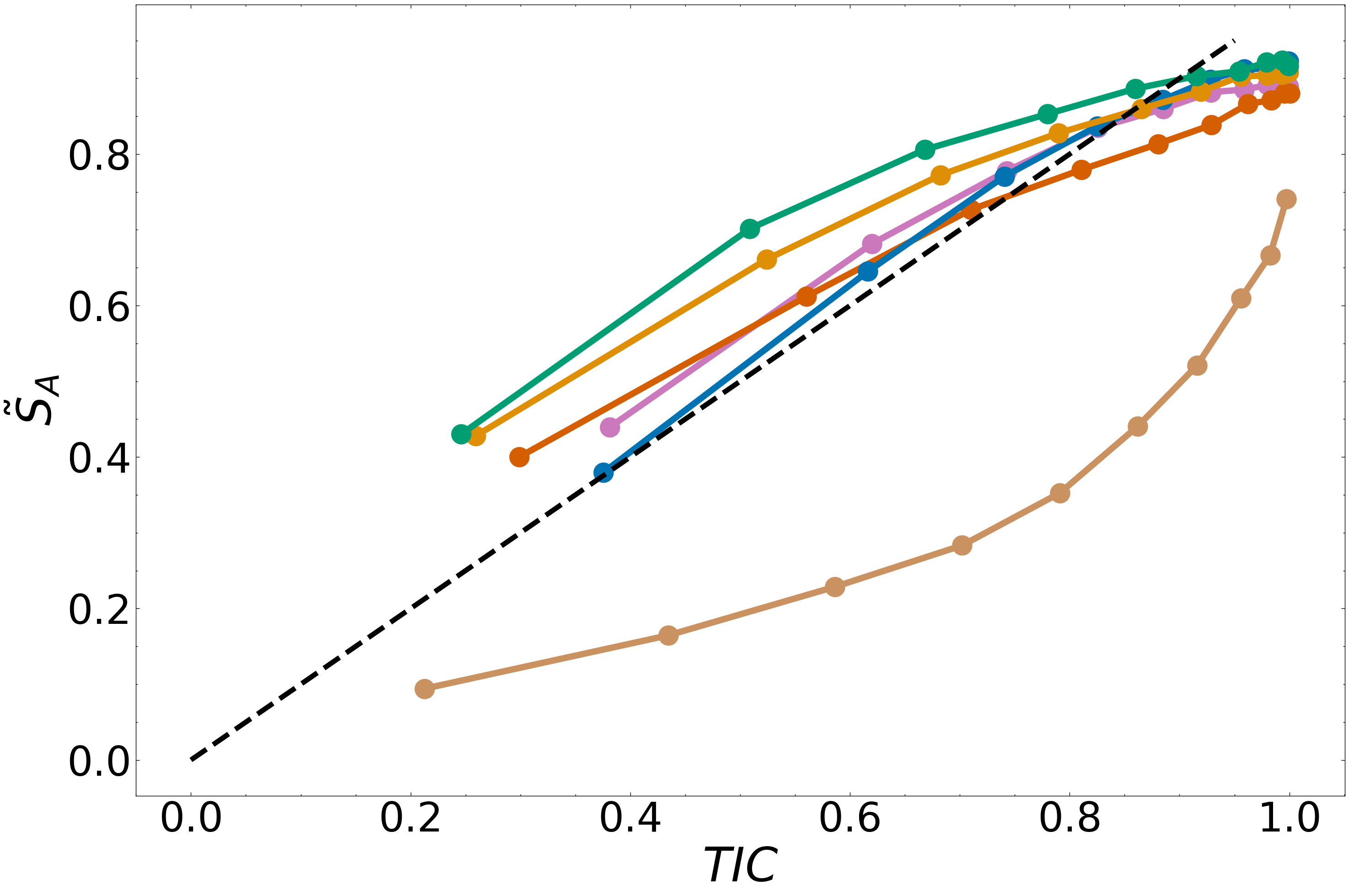}
         \vspace{-0.5cm}
         \caption{Transformer}
     \end{subfigure}
     \hfill
      \begin{subfigure}[b]{0.4\textwidth}
         \centering
         \includegraphics[width=\textwidth]{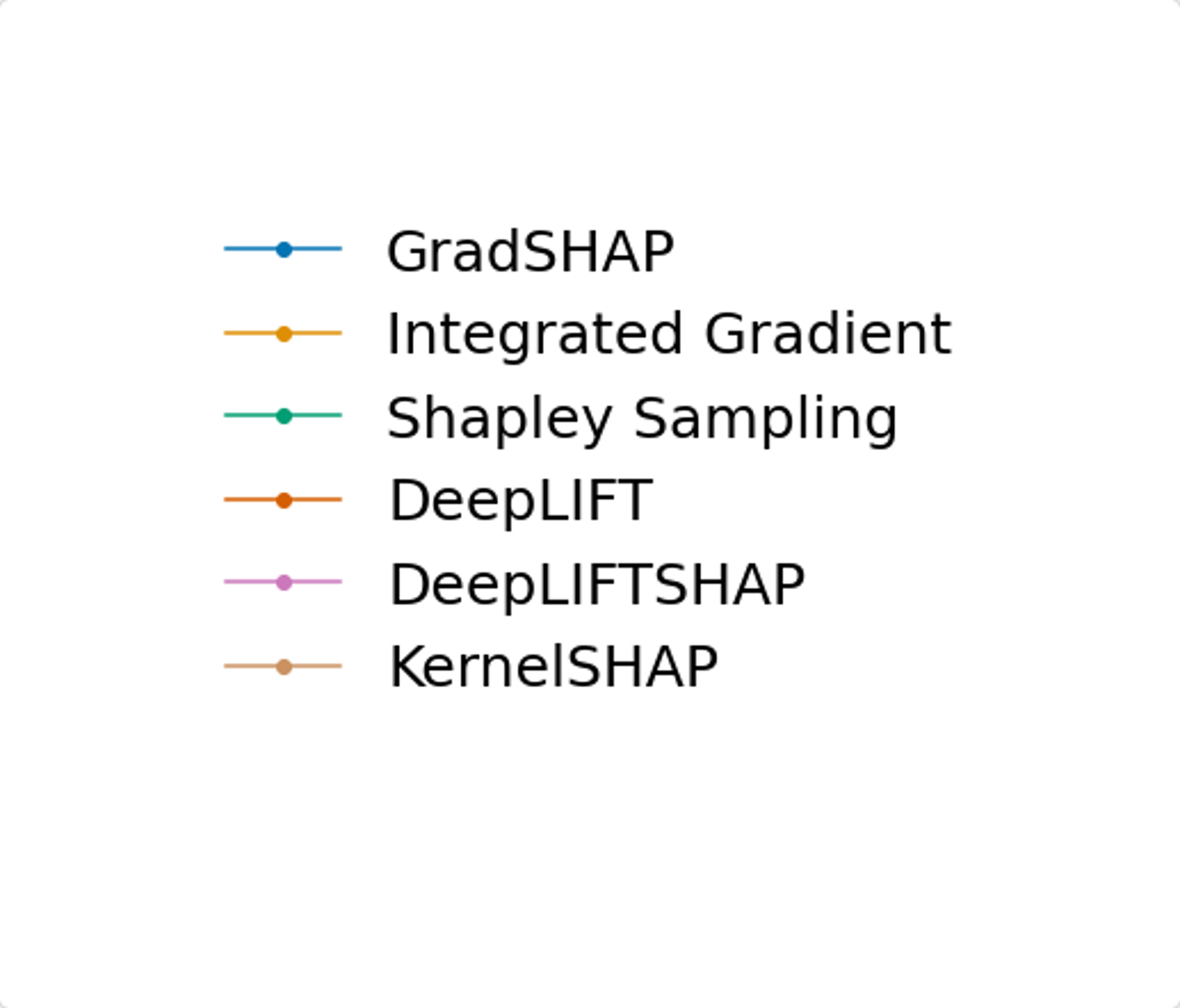}
         \caption*{}
     \end{subfigure}
     \caption{\textbf{$\tilde{S}_A$ as a function of the TIC index for the six interpretability methods considered using the ECG dataset.} Results depicted for (a) Bi-LSTM, (b) CNN and (C) Transformer.}
        \label{fig:rel-attribution_ecg}
     \end{figure}

Table~\ref{table:metrics_interp_normal} shows the two metrics, $AUC\tilde{S}_{\textrm{top}}$ and $F1\tilde{S}$, for all the datasets considered as well as for all interpretability methods and neural-network architectures. 
In table~\ref{table:metrics_interp_normal}, we write in bold the best performing interpretability method for a given architecture and dataset, while in italic the worst performing one. 
In addition, the observed drop in accuracy for samples being progressively corrupted is presented in extended data figure~\ref{fig:accuracy_synthetic} for the synthetic dataset, and \ref{fig:accuracy_ecg} for the ECG dataset while results for the FordA datasets are presented in Supplementary information, section 1.2.
\begin{table}[H]
\centering
\caption{\textbf{$AUC\tilde{S}_{\textrm{top}}$ and $F1\tilde{S}$ metrics for all datasets, interpretability methods and neural networks considered in this work}. \textbf{Bold} indicates the best performing interpretability method for the given metric and dataset and \textit{italic} the worst performing one. IG stands for Integrated Gradients.}
\label{table:metrics_interp_normal}
\begin{tabular}{@{}ll|ccc|ccc@{}}
\toprule
& & \multicolumn{3}{c}{$AUC\tilde{S}_{\textrm{top}}$} & \multicolumn{3}{c}{$F1\tilde{S}$} \\
\cline{3-8}
 Network &   Method    & Synthetic  & Ford A  & ECG   & Synthetic   & Ford A & ECG  \\ 
\cline{1-8}
\multirow{7}{*}{\textbf{Bi-LSTM}} 
& DeepLift      & 0.381 & 0.248 & 0.165 & 0.165 & 0.105 & 0.068 \\
& DeepLiftShap  & 0.439 & 0.296 & 0.156 & 0.182 & 0.108 & 0.061 \\
& GradShap      & 0.392 & 0.283 & 0.262 & 0.172 & 0.126 & 0.110 \\
& IG            & 0.480 & 0.364 & 0.326 & 0.196 & 0.130 & 0.131 \\
& KernelShap    & {\it0.291} & {\it0.207} & {\it0.099} & {\it0.131} & {\it0.084} & {\it0.028} \\
& Shapley       & {\bf0.554} & {\bf0.380} & {\bf0.348} & {\bf0.210} & {\bf0.173} & {\bf0.139} \\ 
\cline{2-8}
&Random         & 0.302 & 0.181 & 0.088 &  &  &  \\ 
\cline{2-8}
\midrule
\multirow{7}{*}{\textbf{CNN}} 
& DeepLift      & 0.626 & 0.483 & 0.280      & 0.245 & 0.206 & 0.122 \\
& DeepLiftShap  & 0.717 & 0.487 & {\bf0.465} & {\bf0.284} & {\bf0.259} & {\bf0.200} \\
& GradShap      & 0.673 & 0.419 & 0.357      & 0.252 & 0.171 & 0.156 \\
& IG            & 0.659 & 0.485 & 0.313      & 0.246 & 0.178 & 0.134 \\
& KernelShap    & {\it0.335} & {\it0.247}  & {\it0.025} & {\it0.155} & {\it0.102} &\it{-0.006} \\
& Shapley       & {\bf0.757} & {\bf0.499} &0.341 & 0.276 & 0.194 & 0.142 \\ 
\cline{2-8}
& Random        & 0.312 & 0.214 & 0.047 & & & \\ 
\cline{2-8}
\midrule
\multirow{7}{*}{\textbf{Transformer}} 
& DeepLift      & 0.652 & 0.309 & 0.441 & 0.243 & {\it0.096} & 0.168 \\
& DeepLiftShap  & 0.800 & 0.330 & 0.539 & 0.277 & 0.107 & 0.217 \\
& GradShap      & 0.943 & 0.382 & 0.584 & 0.301 & 0.141 & 0.222 \\
& IG            & 0.929 & 0.438 & 0.581 & 0.301 & 0.170 & 0.215 \\
& KernelShap    & {\it0.362} & {\it0.272} & {\it0.150} & {\it0.156} & 0.108 & {\it0.051} \\
& Shapley       & {\bf0.943} & {\bf0.650} & {\bf0.619} & {\bf0.303} & {\bf0.245} & {\bf0.228} \\ 
\cline{2-8}
& Random        & 0.407 & 0.248 & 0.156 & & & \\ 
\cline{2-8}
\bottomrule
\end{tabular}
\end{table}
The \textit{relevance identification} evaluated through the two metrics presented above focuses on assessing how the relevance produced by interpretability methods allows ordering time steps to extract the most, or least, relevant time steps for a model. 
Another important aspect of interpretability methods is their capacity to estimate the relative effect of a given time step on the final prediction. 
We call this aspect \textit{relevance attribution}. 
Developing on properties of the interpretability methods included in this work, 
the \textit{relevance attribution} of interpretability methods are evaluated qualitatively using curves of the adjusted normalized change in score $\tilde{S}_A$, defined in Methods section~\ref{sec:methods_evaluation} in equation~\eqref{eq:mod-S_A}, versus the time series information content (TIC) index.
The TIC index measures the proportion of positive relevance contained in the corrupted portions of the time series. 
Figure~\ref{fig:rel-attribution_ecg} shows $\tilde{S}_A$ as a function of the TIC index measured on the ECG dataset. 
This figure allows a qualitative evaluation of the \textit{relevance attribution} performance of interpretability methods. 
If a curve is above the theoretical unit linear slope (depicted as the black line), the interpretability method under-estimates the influence of the corrupted time steps with regard to their effect on the model's prediction. 
The opposite is true if the curve stands below the unit slope. The evaluation of the \textit{relevance attribution} can therefore be seen as a measure of how well calibrated an interpretability method is in terms of the relevance it assigns to the different time steps with respect to their importance for the model to make its predictions. Similar figures for the synthetic dataset and the Ford A dataset are presented respectively in extended data figure~\ref{fig:rel-attribution_synthetic} and Supplementary information, section 1.3.
\section{Discussion and conclusion}
\label{sec:discussion}

This paper presents a new evaluation method and a set of evaluation metrics for post-hoc interpretability to answer the question posed in the introduction: \textit{what method produced an interpretability map closer to the one actually adopted by the neural network to make its prediction?} The two new metrics, namely $AUC\tilde{S}_{\textrm{top}}$ and $F1\tilde{S}$, allow quantifying the \textit{relevance identification} performance of an interpretability method, and can be used to e.g., rank interpretability methods. 
These two metrics agree in identifying Shapley Value Sampling as the top performing method (see table~\ref{table:metrics_interp_normal}).

Focusing on the $AUC\tilde{S}_{\textrm{top}}$ presented in table ~\ref{table:metrics_interp_normal}, Shapley Value Sampling consistently outperforms the other interpretability methods across the different datasets and architectures (except for the CNN trained on the ECG dataset). The $AUC\tilde{S}_{\textrm{top}}$ metric reflects the capacity of Shapley Value Sampling to extract the most important time steps for a model prediction. 
Shapley Value Sampling is however the most computing-intensive interpretability method among the ones tested in this paper. It can therefore be convenient to look for alternatives. These alternatives depend on the type of architecture selected. Integrated Gradients is the second best interpretability method for Bi-LSTM networks while DeepLiftShap is ranked second for CNN. The results are slightly less clear for Transformer network where GradShap and Integrated Gradients have very similar performance.

In addition to $AUC\tilde{S}_{\mathrm{top}}$, the $F1\tilde{S}$ metric measures the ability of different interpretability methods not only to select the most important time steps but also the least important ones. 
The rankings produced using the two metrics are consistent with one another for both transformer and Bi-LSTM while favoring DeepLiftShap for CNN architecture.

In order to obtain reliable results for $AUC\tilde{S}_{\mathrm{top}}$ and $F1\tilde{S}$, we addressed the issues identified in the literature. 
In particular, we evaluated the interpretability methods avoiding human judgement, and we did not retrain the model, while avoiding a distribution shift between the training set and the occluded set used to evaluate the interpretability methods. 
Indeed, the distribution shift is one of the main concerns found in the interpretability literature. 
The method proposed in this paper, described in detail in Method section~\ref{sec:model_training}, addresses this issue. 
Hence, the drop in score observed in figure~\ref{fig:syntethic-dataset-identification_transformer} as the samples are progressively corrupted cannot be attributed to a distribution shift. 
We also note that a larger drop in score is systematically observed when corrupting the most relevant time steps (identified by the interpretability method) as compared to corrupting a random selection of time steps (black line in figure~\ref{fig:syntethic-dataset-identification_transformer}), as expected. 
To the authors' knowledge, the approach presented in this paper is the first that allows quantitatively evaluating interpretability methods without retraining, while avoiding a distribution shift between training and evaluation set. 

The two metrics $AUC\tilde{S}_{\mathrm{top}}$ and $F1\tilde{S}$, along with the majority of the literature on interpretability methods evaluation focuses on relevance identification (i.e., ranking time steps according to their importance). 
In this work, we also build a first step towards evaluating \textit{relevance attribution}. 
This evaluates how the relevance reflects the relative importance of each time step compared to the others. 
The attribution is qualitatively evaluated using the $\tilde{S}_{A}-\text{TIC}$ curves (figure~\ref{fig:rel-attribution_ecg}). These curves provide an understanding of the ability of an interpretability method to correctly weigh relevance and are compared to a newly derived theoretical estimation (derived in Supplementary information, section 4). This theoretical estimate is depicted in figure~\ref{fig:rel-attribution_ecg} as a black dashed line. 
The \textit{relevance attribution} performance is consistently different between the different neural networks tested and it also varies between datasets. 
The common denominator is the inability of the interpretability methods to follow the theoretical estimate. 
This indicates that the relevance attributed to each time step does not reflect the relative importance of this time step in the classification task. 
Instead, the attributed relevance acts more as a ranking of the most important time steps among themselves. 
For example, a point with a relevance of 0.1 for a total classification score of 1 might not necessarily account for 10\% of the final prediction, but will be more important than a point with a relevance equal to 0.05. 
Albeit qualitative, these curves may be used to assess visually whether an interpretability method provides a balanced relevance.

As part of this work, we also provide a new synthetic dataset that can be used for evaluating interpretability methods.
The new dataset forces the neural network to learning time dependencies as opposed to learning static information, and the discriminative portions of the time series are known a priori. 
Additionally, the dataset is multivariate by construction, that is a desirable property especially when trying to mimic real-world (i.e., non-synthetic) datasets. 
The performance of the interpretability methods on the new synthetic dataset are consistent with the performance obtained on the two real-world datasets tested as part of this work, namely Ford A and ECG.
Hence, the designed dataset acts as a good proxy for real-world classification tasks with two convenient properties: its complexity and properties are tuneable, its generation is lightweight. 
As such, it can complement real-world datasets for a range of different research objectives within the context of evaluating post-hoc interpretability methods in time-series classification, given its known multivariate and time-dependent discriminative properties.

Finally, we assessed the usefulness of interpretability methods validated with our evaluation framework in an operational setting. In particular, we used the ECG clinical dataset, as it provides a good example of how interpretability can be used once the interpretability methods have been evaluated. The use of clinical data was favored because the healthcare sector is likely to become highly regulated, hence require accurate interpretability of AI technologies~\cite{ai_act}. To this end, we interacted with clinicians to understand a common disease that is representative in the context of ECGs, and that can be of interest to them. This turned out to be the well studied cardiac disease, right bundle branch block (RBBB).
In the classification task presented in figure~\ref{fig:differences_interp} for the ECG data, the neural networks were trained to classify RBBB. Shapley Value is able to pinpoint a specific and compact region of interest in the time series, while the other methods provide interpretability maps that are less compact (in the case of KernelShap, a sparse map, without a clear region).
Interestingly, the feature highlighted by the Shapley Value relevance map corresponds to one of the morphological features  cardiologists look at to diagnose the disease of interest \cite{SURAWICZ2009976}. The interpretability method also shows that the trained model relies almost entirely on a single lead to predict the disease in question, namely RBBB. However, other diagnostic criteria focusing on different leads are commonly used. This type of analysis provides practical insights to understand how trained models will perform in an applied operational setting, and may help identifying possible biases, spurious correlations and potential corrective actions. Of interest, going forward, is understanding how these analyses could be implemented for e.g., regulatory purposes, and deployed as part of AI-based technologies in new high-risk applications.

\setcounter{section}{0}
\renewcommand*{\theHsection}{chX.\the\value{section}}
\section*{Methods}
\label{sec:methods}

\section{Tackling distribution shift}
\label{sec:model_training}
A longtime issue when evaluating interpretability methods has been the shift in distribution between the training dataset and the corrupted dataset used for the evaluation. 
Indeed interpretability methods have been frequently evaluated comparing the drop in score when the most relevant time steps are corrupted with the score on the initial sample. 
The \textit{ROAR} approach was proposed to address this issue \cite{hooker_benchmark_2019}, however retraining the model of interest comes with its own drawbacks, that were discussed in the introduction of this paper. 
The training method presented next aims to address this issue, thereby maintaining a constant distribution between the training dataset and the one used to evaluate the interpretability method. 
In order to achieve this task, the models presented in this paper were trained with a random perturbation applied to the time series. 
This method was inspired by data-augmentation strategies commonly used when training models for image classification, object detection, and other image-based tasks. 
On these tasks, random cropping has been shown to improve the classification performance of the developed model as well as their robustness~\cite{NEURIPS2020_cubuk}. 
In our case, the aim of the perturbations in the training set is not related to improving the performance of the model, but rather to maintain an identical distribution between the training dataset and the corrupted samples used to evaluate the interpretability methods.

Similarly to random cropping applied to images, part of the times series is corrupted by substituting the initial time steps with points drawn from a normal distribution $\sim \mathcal{N}(0,1)$. This distribution follows the normalisation applied as a preprocessing step to the samples.
In a similar fashion as \textit{DropBlock}~\cite{ghiasi2018dropblock}, consecutive time steps, or blocks, are corrupted. 
The augmentation is applied per batch with the overall fraction of the time series being corrupted ($\gamma$) and the size of the blocks ($\beta$) being sampled from the following uniform distributions: \begin{align}
    \gamma & \sim \mathcal{U}(0, 0.8)  \\
    \beta & \sim  \mathcal{U}(1,7) 
\end{align}

Given the method described above, when specific time steps are corrupted to evaluate the interpretability methods, the distribution is retained with the samples used when training the model. $\beta$ is chosen to reflect the range of consecutive time steps above the median positive relevance empirically observed over the used datasets.$\gamma$ was also empirically chosen to cover most of the samples where the positive relevance is rarely assigned to more than $80\%$ of the total number of time steps in a given sample. The change in score observed when corrupting time steps can therefore not be attributed to a shift in the distribution, and hence fully reflect a loss of information for the model as measured by the interpretability methods. 
To the authors' knowledge, this is the first time that an approach that addresses the distribution shift in the evaluation of interpretability methods has been proposed without requiring retraining the model. 
The latter point is important as it is not possible to assert that the retrained model uses the same time steps as the initial one which the interpretability methods aim to explain.

\section{Novel approach for evaluating post-hoc interpretability methods}
\label{sec:interpretability}
The time-series classification task considered in this paper can be formalized as follows, where the symbols and notation adopted are also reported in table~\ref{table:variable_summary}. 

Given a trained neural-network model $f$, we aim to map a set of features $\mathbf{X} \in \mathbb{R}^{M \times T}$ to a labelled target $\text{C} \in \mathbb{N}^{N_{c}}$, for each sample $i$ contained in a given dataset $\mathcal{D}_i = [\mathbf{X}, \text{C}]_i \;\; \text{for}\; i = 1, \dots, J$, where $M$ is the number of features, $T$ is the number of time steps per feature, $N_{c}$ is the  number of labels, and $J$ is the total number of samples available. 
Typically, a final dense layer will produce logits that are then fed to a softmax layer to output the probability of sample $i$ to belong to a given class C $\in \mathbb{N}^{N_c}$. 

To assess quantitatively time-series interpretability methods, we developed novel metrics or indices to evaluate how closely an interpretability method reflects the representation learned by the model of interest.
Interpretability methods produce an attribution scheme $\mathcal{A}$ which assigns relevance $\mathbf{R}$ to the input $\mathbf{X}$ for a specific class $c \in C$ such that $\mathcal{A}_{c}: \mathbf{X} \rightarrow \left \{\mathbf{R} \in \mathbb{R}^{M \times T } \right \}$, where $\mathbf{X} = \left(x_{m,t}\right)$ and  $\mathbf{R} = \left(r_{m,t}\right)$, with $m$ and $t$ being the indices associated to the number of features $M$ and to the number of time steps $T$, respectively. For simplicity, the class that the attribution scheme aims to explain is dropped in the rest of the paper and the attribution scheme is denoted by $\mathcal{A}$. 
The new metrics are built on the relevance that an interpretability method provides along the time series. 
The relevance can be positive or negative (except for some interpretability methods, where it is only positive -- see e.g., the saliency method~\cite{simonyan_deep_2014}). 
A positive relevance means that the neural network is using that portion of the time series to make its prediction. 
A negative relevance indicates that the neural network sees the portion of the time series as going against its prediction. 
As we are interested in how the network is using data to making its predictions, we use the positive relevance to build the new metrics. Logits have often been used as the input for interpretability methods~\cite{simonyan_deep_2014,selvaraju2017grad}. 
However, Srinivas et al.~\cite{srinivas2020rethinking} demonstrate that pre-softmax outputs are related to a generative model which is uninformative of the discriminative model used for the classification task. 
In this sense, the evaluation of the interpretability methods for the rest of the paper are produced with the post-softmax models' output as well as evaluated with changes in these outputs when corrupting samples, denoted by $S: \mathbb{R}^{M \times T} \rightarrow \left[0,1\right]^{N_{c}}$.

In this work, we aim to build a framework to evaluating  the performance of interpretability methods. To this end, we chose six interpretability methods that capture a broad range of methods available, while keeping the problem computationally tractable. These are: i) DeepLift~\cite{deeplift}, ii) GradShap~\cite{NIPS2017_8a20a862}, iii) Integrated Gradients~\cite{ig_method}, iv) KernelShap~\cite{NIPS2017_8a20a862}, v) DeepListShap~\cite{NIPS2017_8a20a862}, vi) Shapley Value Sampling~\cite{CASTRO20091726}. 
Their implementation uses the Captum library~\cite{kokhlikyan2020captum}.

\subsection{Relevance identification and attribution}
\label{sec:methods_evaluation}
For assessing interpretability methods' performance, we need to tackle two aspects, \textit{relevance identification} and \textit{relevance attribution}. 
We detail these two concepts along the methods developed to measure them next.

\begin{description}
\item \textbf{Relevance identification.} 
The concept behind relevance identification is that interpretability methods should correctly identify and order according to their relevance the set of points (in our case time steps) used by the model to make its predictions. 
Extending the assumption formulated by Shah et al.~\cite{shah2021input}, time steps with larger relevance are more relevant for the model to make a prediction than the ones with smaller relevance. 
The relevance produced by an attribution scheme can be used to create an ordering that ranks feature $m$ at time step $t$, namely $x_{m,t}$, from a sample $\mathbf{X}$ according to its importance for the model's prediction. 
It is important to remember that here we only focus on the positive relevance. 
The ordering can then be used to define  $\mathbf{\bar{X}}^{\text{top}}_k$ and $\mathbf{\bar{X}}^{\text{bottom}}_k$ that represent the samples with top-$k$ and bottom-$k$ time steps corrupted, respectively. These are ordered using the assigned relevance, corrupted as follows:  
\begin{align}
    \mathbf{\bar{X}}^{\text{top}}_k  &= \begin{cases}
        x_{m,t} & \text{if } r_{m,t} < Q_{R^+}(1-k) \\
    \mathcal{N}(0,1)              & \text{otherwise} 
\end{cases} \\
\mathbf{\bar{X}}^{\text{bottom}}_k  &= \begin{cases}
        x_{m,t} & \text{if } r_{m,t} > Q_{R^+}(k)  \\
    \mathcal{N}(0,1)            & \text{otherwise},
\end{cases}
\end{align}
where $Q_{R^+}(p)$ denotes the $p$-quantile (with $p=1-k$, for $\mathbf{\bar{X}}^{\text{top}}_k$, and $p = k$, for $\mathbf{\bar{X}}^{\text{bottom}}_k$) of the positive relevance ${R}^+$, measured over sample $\mathbf{X}$ using attribution scheme $\mathcal{A}$ with ${R}^+ = \left \{r_{m,t} \vert r_{m,t}>0 \right \}$. 
The rest of the analysis is performed for the following set of top-$k$ and bottom-$k$ percentage of time steps with positive relevance: $k \in [0.05, 0.15,0.25,...,0.95,1]$ 

In the general case of a modified sample $\mathbf{\bar{X}}$ where $\bar{N}$ points along the time series are corrupted we can define the normalized difference in score: 
\begin{equation}
    \tilde{S} (\mathbf{\bar{X}}) = \frac{S(\mathbf{X})- S(\mathbf{\bar{X}})}{S(\mathbf{X})}
    \label{eq.s_tilde}
\end{equation}

It is possible to build $\tilde{S} \; \text{versus}\; \tilde{N}$ curves for top-$k$  or bottom-$k$ points (or time steps) corrupted, where $\tilde{N} = \frac{\bar{N}}{N}$ is the fraction of points removed with respect to the total number of time steps $N = M \times T $ present in the time series. The area under the $\tilde{S}-\tilde{N}$ curve is denoted as: 
\begin{equation}
    AUC\tilde{S} = \int_{0}^{1} \tilde{S} \;\text{d}\tilde{N}
    \label{eq:aucs}
\end{equation}
Using equation~\eqref{eq:aucs}, we can define the two following metrics: $AUC\tilde{S}_{\textrm{top}}$ and $F1\tilde{S}$. 
The $AUC\tilde{S}_{\textrm{top}}$ metric aims to evaluate the ability of an interpretability method to recover the most important time steps for a model's prediction. 
In this sense, the area under the drop in score when the top-$k$ time steps are progressively corrupted should be maximized. 
In order to normalize for the interpretability methods assigning a different number of time steps with positive relevance, the $AUC\tilde{S}_{\textrm{top}}$ is measured on a modified $AUC\tilde{S}$ curve. 
This modified curve is created by adding an extra point with coordinates $\left(\tilde{N} = 1; \tilde{S} = \tilde{S}(\mathbf{\bar{X}}^{\text{top}}_{k=1})\right)$. 
Adding this point allows favoring interpretability methods which are able to achieve a large drop in score with a minimal number of time steps assigned with positive relevance.

The $F1\tilde{S}$ aims to build an harmonic mean between the ability of an attribution scheme to rank correctly the time steps with the highest, and smallest relevance, respectively. Corrupting time steps with  high relevance should result in a significant drop in score (model's outputs are significantly affected). Corrupting time steps with small relevance should result in a negligible drop in score (model's outputs are negligibly affected). 
The best attribution scheme should have maximized $AUC\tilde{S}_{\mathrm{top}}$ for top-$k$ points corruptions, and minimized $AUC\tilde{S}_{\mathrm{bottom}}$ for bottom-$k$ points corruptions.
Regarding this desired property, we can define the following $F1$ score:
\begin{equation}
    F1\tilde{S} = \frac{AUC\tilde{S}_{\mathrm{top}} \left(1-AUC\tilde{S}_{\mathrm{bottom}}\right)}{AUC\tilde{S}_{\mathrm{top}} + \left(1-AUC\tilde{S}_{\mathrm{bottom}}\right)}
    \label{eq:F1-aucse}
\end{equation}

\item \textbf{Relevance attribution.} 
The idea behind relevance attribution is that the relevance should not only serve to order the time steps but also reflect the individual contribution of each time step  relative to the others towards the model's predicted score. All interpretability methods presented in this work are additive feature attribution methods as defined by Lundberg et al. \cite{NIPS2017_8a20a862}. The produced relevance therefore aims to linearly reflect the effect of each feature on the model's outputs. Focusing on the positive relevance, the difference between the model's prediction on an initial given sample $S(\mathbf{X})$ and a version with the top-$k$ points corrupted $S(\mathbf{\bar{X}}^{\text{top}}_k)$ is dependent on the positive relevance corrupted between the two samples. The proportion of relevance attributed to corrupted time steps to the initial one is summarised with the time information content index defined as: 
\begin{equation}
    \text{TIC}(k) = \frac{\sum\limits_{R_k^+} r_{m,t}}{\sum\limits_{R^+} r_{m,t} + \epsilon}
    \label{eq:tic}
\end{equation} 

where the following sets are defined: \begin{align}
    R^+ &= \left \{ r_{m,t} \vert r_{m,t}>0   \right \} \\
    R^+_k &= \left \{ r_{m,t} \vert r_{m,t} \in R^+ \cap  r_{m,t} \geq Q_R(1-k)  \right \} 
\end{align}

The TIC index reflects the ratio of the relevance attributed to the top-$k$ set of points to the total positive relevance.

Taking as a reference the model's output when all the positive relevance is corrupted we can normalize the change in score as follows: 
\begin{equation}
    \tilde{S}_A(k) =   \frac{S(\mathbf{X})- S(\mathbf{\bar{X}}^{\text{top}}_k)}{S(\mathbf{X})- S(\mathbf{\bar{X}}^{\text{top}}_{k=1})}
    \label{eq:mod-S_A}
\end{equation}
where $S(\mathbf{\bar{X}}^{\text{top}}_{k=1})$ corresponds to the model's output when all time steps with positive relevance of sample $\mathbf{X}$ is corrupted. We name the quantity in equation~\eqref{eq:mod-S_A} adjusted normalized drop in score. Given the linear additivity property of the relevance, the index $\tilde{S}_A(k)$ should be equal to the $\text{TIC}(k)$ index (see Supplementary information, section 4) such that the information ratio (IR) satisfies 
\begin{equation}
   \text{IR} = \frac{\tilde{S}_A(k)}{\text{TIC}(k)} = 1
   \label{eq.IR}
\end{equation}
Given this theoretical approximation, it is possible to evaluate how different interpretability methods over- or under-estimate the role of different time steps in the model's prediction. An IR larger than one will indicate the relevance of the points under the quantile of interest was under-estimated while the opposite is true for an IR smaller than one. An example of $\tilde{S}_A(k)-\text{TIC}(k)$ curve is depicted in figure~\ref{fig:rel-attribution_ecg} where we report the results for every interpretability method considered as well as the theoretical linear line (dashed). 

\end{description}

\section{Datasets}
\label{sec:datasets}
The new interpretability evaluation approach has been applied to a new synthetic dataset created for this work, a standard univariate dataset for anomaly classification and a biomedical dataset based on electrocardiogram (ECG) signals. 
The three datasets are described next.

\subsection{A new synthetic dataset}
\label{sec:methods_synthetic_dataset}
The evaluation of interpretability methods for time series classification has been lacking a dataset where: 1) the discriminative features are known and 2) replicates the complexity of common time series classification tasks with time dependencies across features. 
The developed dataset is inspired by the \textit{BlockMNIST} dataset which is derived from the \textit{MNIST} dataset \cite{shah2021input}. 
Each sample in the dataset is composed of 6 features each with 500 time steps corresponding to a $\Delta t = 2 \text{ms}$. 
Each feature is composed of a sine wave with its amplitude multiplied by 0.5 and frequency $\sim \mathcal{U} (2,5)$ which serves as a random baseline. 
In two of the features, picked randomly for each sample, sine waves with a support of 100 time steps are added to the baseline at a random position in time. 

The respective frequency of these sine waves, $f_1$ and $f_2$ are drawn from a discrete uniform distribution $\sim \mathcal{U}(10, 50)$.
In the remaining four features, a square wave is included with a probability of 0.5 and frequency $\sim \mathcal{U}(10, 50)$.

The classification task then consists in predicting whether the sum of the two frequencies ($f_1$ and $f_2$) is above or below a given threshold $\tau$.
For the presented task, $\tau$ was set to 60 to balance the classes of the classification target $y$ such that : 
\begin{equation}
    y = \begin{cases}
        0,& \text{if } f_1 + f_2 < \tau \\
    1,              & \text{otherwise}
\end{cases}
\end{equation}
The main idea behind the developed dataset is to force the network to learn temporal dependencies, i.e., the frequency of the sine wave with closed support, as well as dependencies across features, the sum of the frequencies ($f_1$ and $f_2$). An example of a generated sample is presented in figure~\ref{fig:synthetic_data}. The closed support sine waves used to create the classification target are observed in feature 2 and feature 6. We note that the synthetic dataset proposed here can be regarded as a family of datasets, as the number of features, length of time series, class imbalance and discriminative features are tunable.

\subsection{Ford A}\label{sec:methods_forda_dataset}
The Ford A dataset is part of the UCR Time Series classification archive which aim to group different dataset for time series classification (TSC) \cite{bagnall16bakeoff}. The Ford A is a univariate and binary classification task. The data are originating from an automotive subsystem and the classification task aims to find samples with a specific anomaly. The dataset comes with a train (n=3601) and test split (n=1320), which was retained in this paper. The dataset is of interest as it has often served as a benchmark for classification algorithms \cite{voice2series} as well as for benchmarking interpretability methods \cite{schlegel2019towards}.

\subsection{ECG dataset}\label{sec:methods_ecg_dataset}
To mimic a real-world classification task, we applied the interpretability framework to an ECG dataset. ECG records the electrical activity of the heart and typically produces 12 signals, corresponding to 12 sensors or leads.  For this task, a subset of the \textit{Classification of 12-lead ECGs: the PhysioNet - Computing in Cardiology Challenge 2020} published under CC Attribution 4.0 License was used. The dataset was narrowed down to the CPSC subset~\cite{Liu2018} which included 6877 ECGs annotated for 9 cardiovascular diseases. As part of these annotations, it was chosen to classify the ECGs for the presence/absence of a Right Bundle Branch Block (RBBB). The dataset includes 5020 cases showing no sign of a RBBB and 1857 cases annotated as carrying a RBBB. RBBB was found to be associated with higher cardiovascular risks as well as mortality~\cite{bussink_rbbb}. 

The data were first denoised using different techniques for low and high-frequency artifacts.  The baseline wander as well as low-frequencies artifacts were first removed by performing Empirical mode decomposition (EMD). The instantaneous frequency is  computed and averaged across each intrinsic mode resulting from EMD to obtain an average frequency of the modes. Modes with an average frequency below 0.7 Hz are then discarded and the signal is reconstructed with the remaining modes. The threshold is a parameter based on the literature where the thresholds range between 0.5 and 1 Hz \cite{thakor_applications_1991, van_alste_removal_1985, van_alste_ecg_1986}. Given the difficulty to separate High-Frequency (HF)  noises using EMD, power-line and others HF noises are removed by thresholding the wavelet transform (WT) coefficients using the "universal threshold" \cite{donoho_ideal_1994}.

In order to obtain an average beat, the R-peaks of each ECG are extracted using the BioSPPy library \cite{biosppy}. The beats centered around the R-peaks are then extracted from the ECG by taking  0.35 before and 0.55 seconds after the R-peak. The mean of the extracted beat is then computed to obtain an "average" beat of each lead. An example of the initial signal along the transformed one for a subset of the 12 leads is presented in extended data figures~\ref{fig:unprocessed_ECG} and \ref{fig:processed_ECG}. In each lead, the average beat was computed. The resulting modified 12 leads were used to train the model.

\section{Baseline for interpretability methods}

The interpretability methods used as part of this work require setting an uninformative baseline as a reference. Most methods (Integrated Gradients, DeepLift, Shapley Value Sampling, KernelShap) require to set a single sample as the baseline. For those methods, the baseline was set as the mean taken across samples for each time step. GradShap and DeepLiftShap uses a distribution of baseline and this baseline was constructed by taking 50 random samples from the test set.

\section*{Data availability} 
All datasets and trained models used in this paper have been made available on Zenodo (\url{https://zenodo.org/record/7534770#.Y8lkkXbMI2w}). The ECG dataset is based on the public dataset released as part of \textit{The PhysioNet/Computing in Cardiology Challenge 2020} available under the following DOI: \url{https://doi.org/10.13026/f4ab-0814}. The Ford A dataset comes from the \textit{UEA \& UCR Time Series Classification Repository} \cite{bagnall16bakeoff}. The synthetic dataset used as part of this study can be generated using the code shared on github: \url{https://github.com/hturbe/InterpretTime}.

\section*{Code availability} 
The code used to perform the analysis is available at \url{https://github.com/hturbe/InterpretTime}.

\section*{Acknowledgments}
Gianmarco Mengaldo acknowledges NUS support through grant R-265-000-A36-133. The authors also thank Dr.\ Adriano Gualandi for fruitful discussions and precious feedback he provided as part of this research. We thank the anonymous reviewers for their insightful comments, that helped significantly improve the manuscript.

\section*{Author Contributions Statement}
H.T. conceived the initial research idea with input from all of the authors. M.B., G.M. and H.T performed the experiments and data analysis. M.B., G.M. and H.T wrote the manuscript with input from all of the authors. C.L. and G.M. supervised the project.

\bibliography{main}
\bibliographystyle{iclr2024_conference}

\setcounter{section}{0}
\setcounter{figure}{0}
\renewcommand{\figurename}{Extended data figure}

\setcounter{table}{0}
\renewcommand{\tablename}{Extended data table}

\newpage
\section*{Extended data}
\begin{figure}[H]
    \centering
    \includegraphics[width=\textwidth]{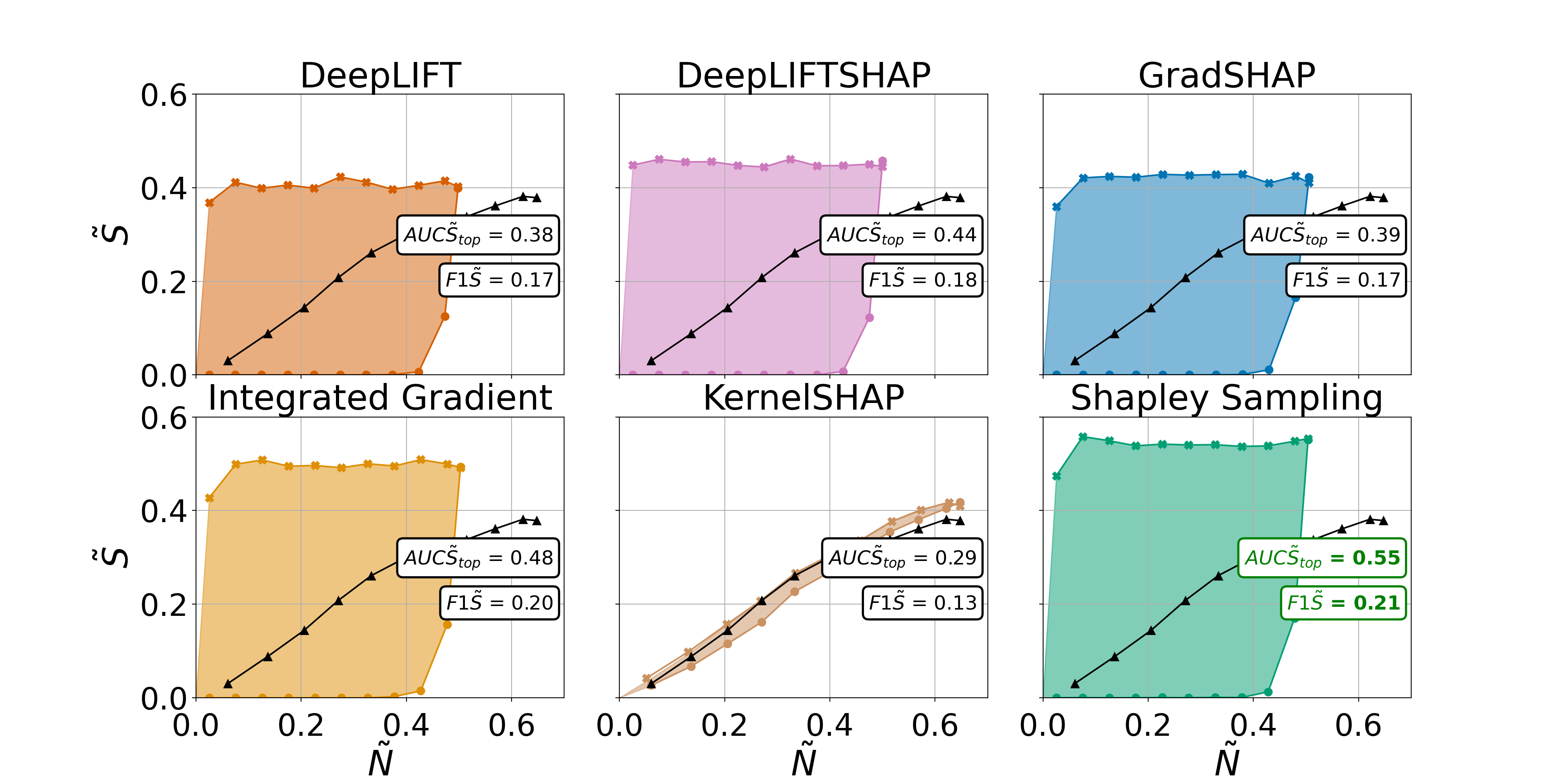}
    \caption{\textbf{$\tilde{S}$ as a function of the ratio of points removed with respect to the total number of time steps in the sample, $\tilde{N}$.} Each subfigure represents one of the six interpretability methods considered for a Bi-LSTM trained on the synthetic dataset.}
    \label{fig:syntethic-dataset-identification_bilstm}
\begin{figure}[H]
         \centering
         \includegraphics[width=\textwidth]{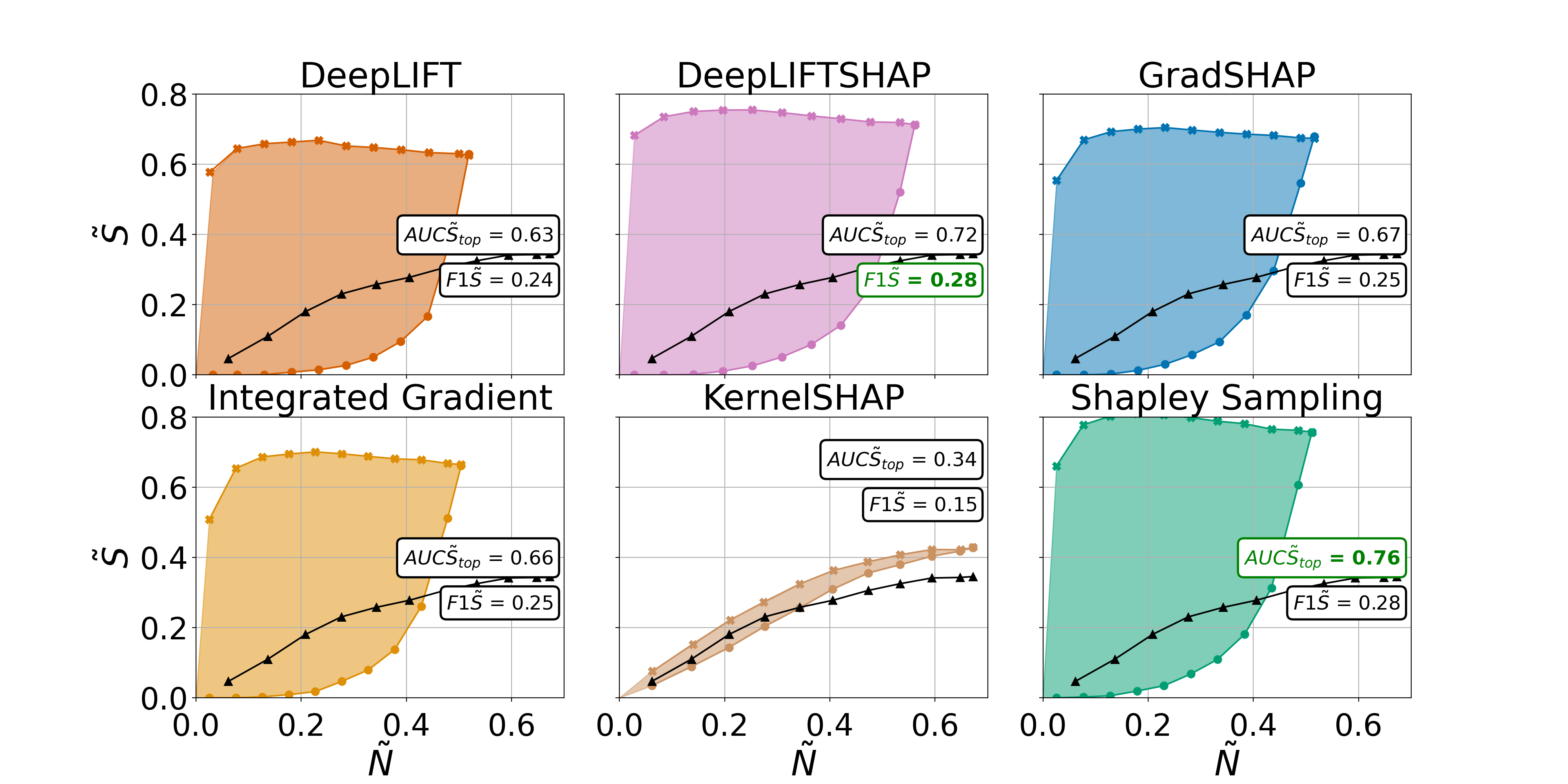}
        \caption{\textbf{$\tilde{S}$ as a function of the ratio of points removed with respect to the total number of time steps in the sample, $\tilde{N}$.} Each subfigure represents one of the six interpretability methods considered for a CNN trained on the synthetic dataset.}
        \label{fig:syntethic-dataset-identification_cnn}
\end{figure}

\end{figure}
\begin{figure}[H]
         \centering
         \includegraphics[width=\textwidth]{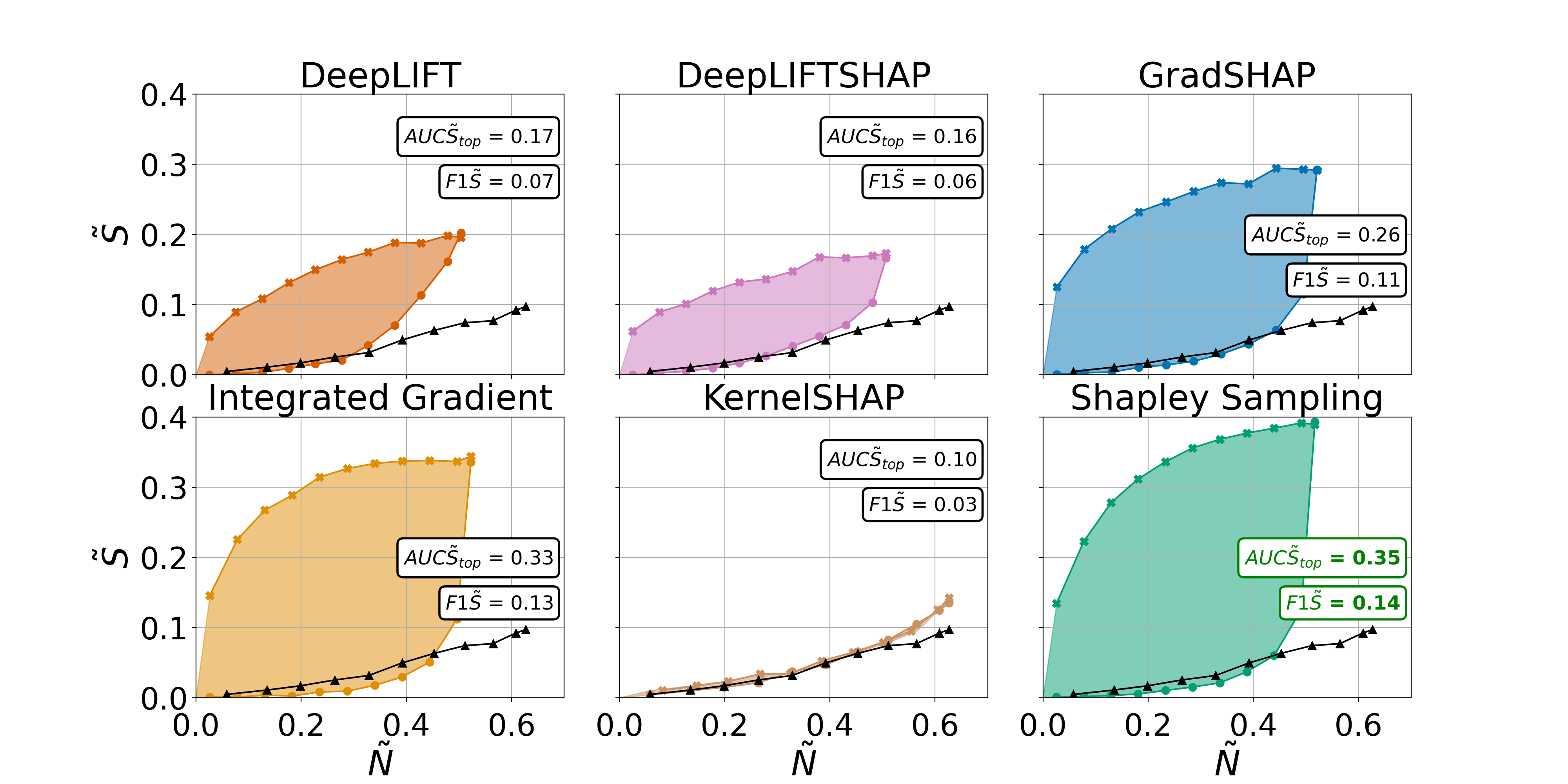 }
        \caption{$\tilde{S}$ as a function of the ratio of points removed with respect to the total number of time steps in the sample, $\tilde{N}$, for the six interpretability methods considered, and for a Bi-LSTM, using the ECG dataset.}
        \label{fig:ecg-dataset-identification_bilstm}
\end{figure}
\begin{figure}[H]
         \centering
         \includegraphics[width=\textwidth]{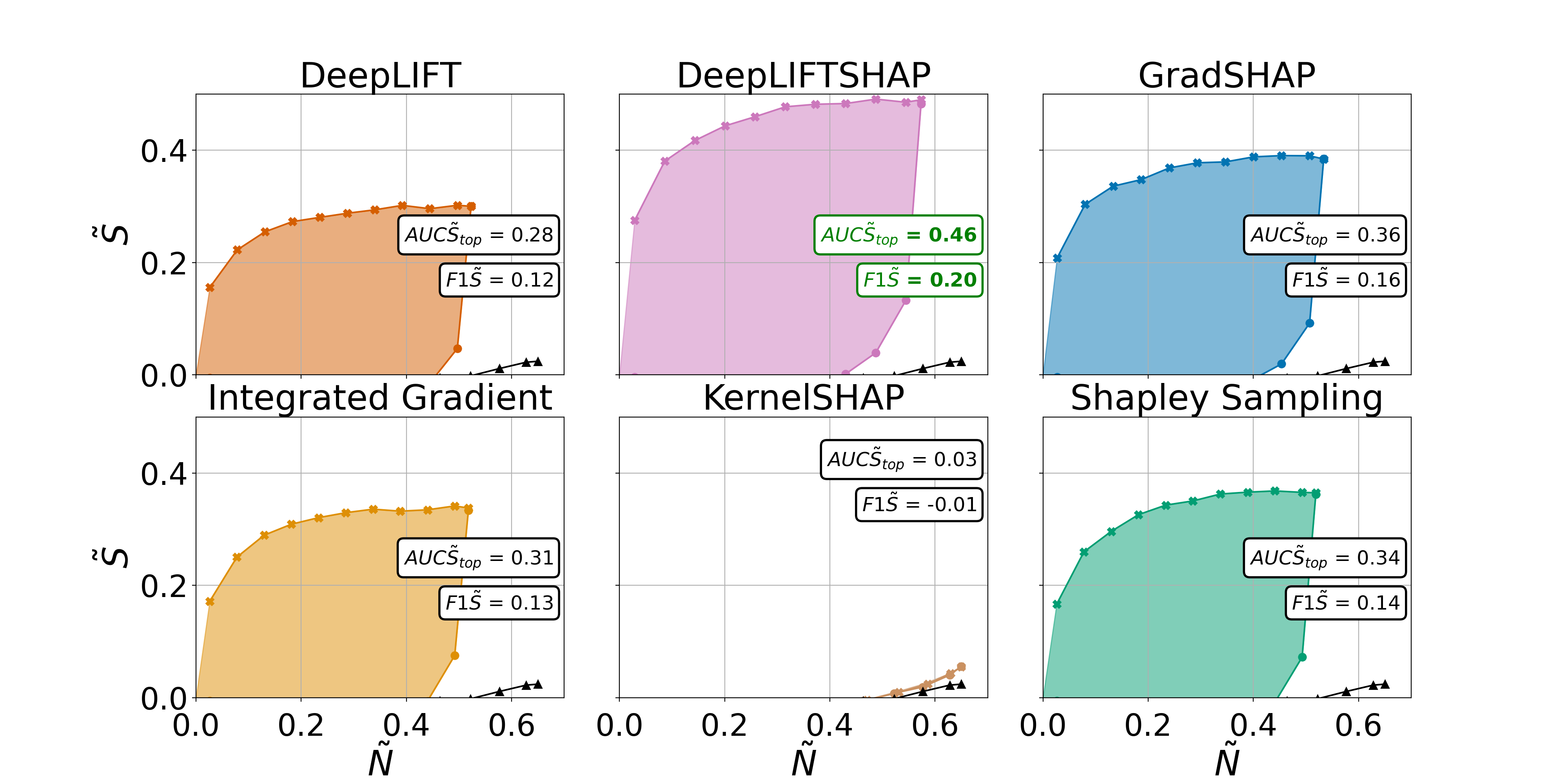}
        \caption{\textbf{$\tilde{S}$ as a function of the ratio of points removed with respect to the total number of time steps in the sample, $\tilde{N}$.} Each subfigure represents one of the six interpretability methods considered for a CNN trained on the ECG dataset.}
        \label{fig:ecg-dataset-identification_cnn}
\end{figure}
\begin{figure}[H]
     \centering
     \includegraphics[width=\textwidth]{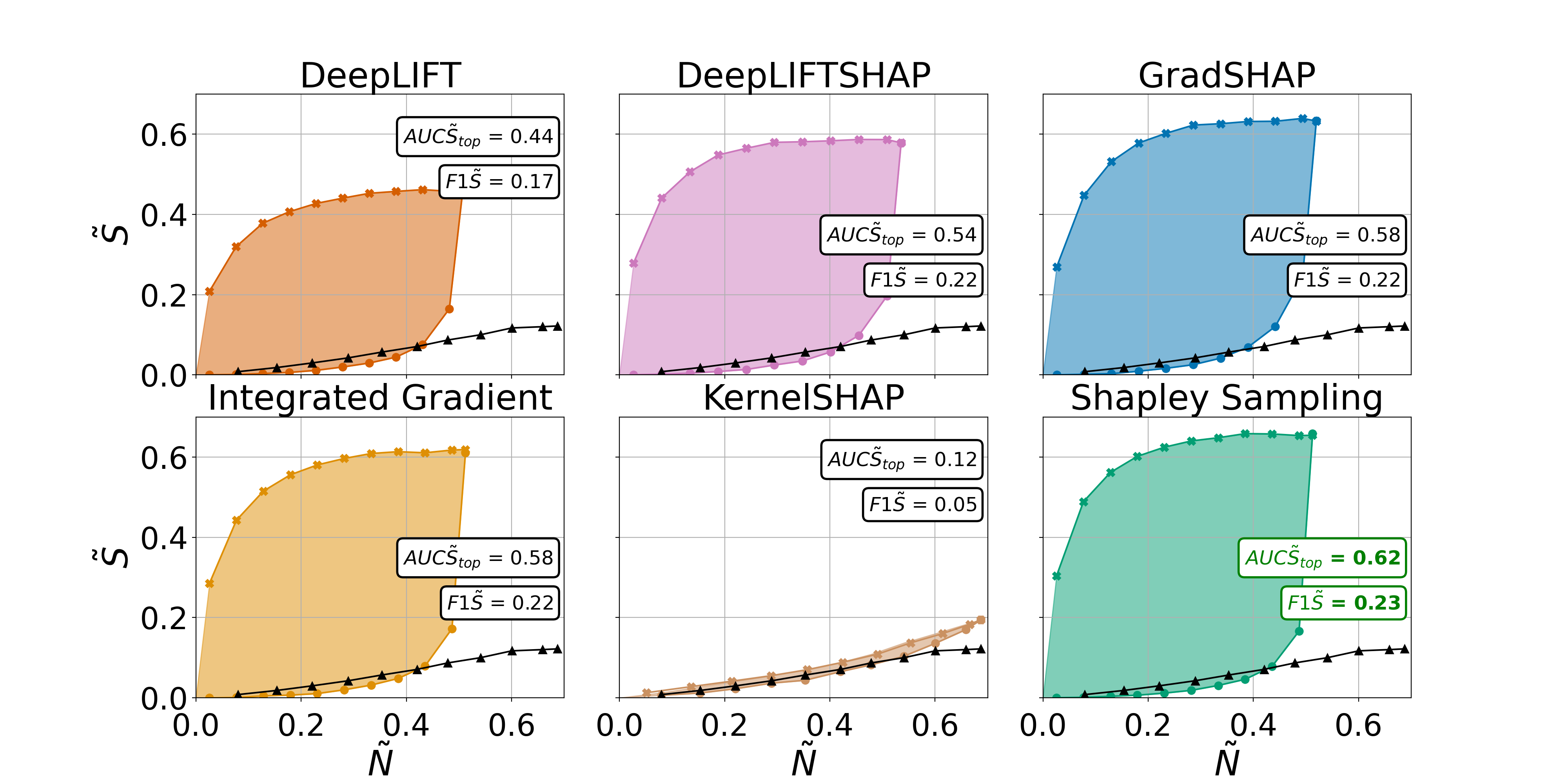}
    \caption{\textbf{$\tilde{S}$ as a function of the ratio of points removed with respect to the total number of time steps in the sample, $\tilde{N}$.} Each subfigure represents one of the six interpretability methods considered for a Transformer trained on the ECG dataset.}
        \label{fig:ecg-dataset-identification_transformer}
\end{figure}
\begin{figure}[H]
     \centering
     \begin{subfigure}[b]{0.49\textwidth}
         \centering
         \includegraphics[width=\textwidth]{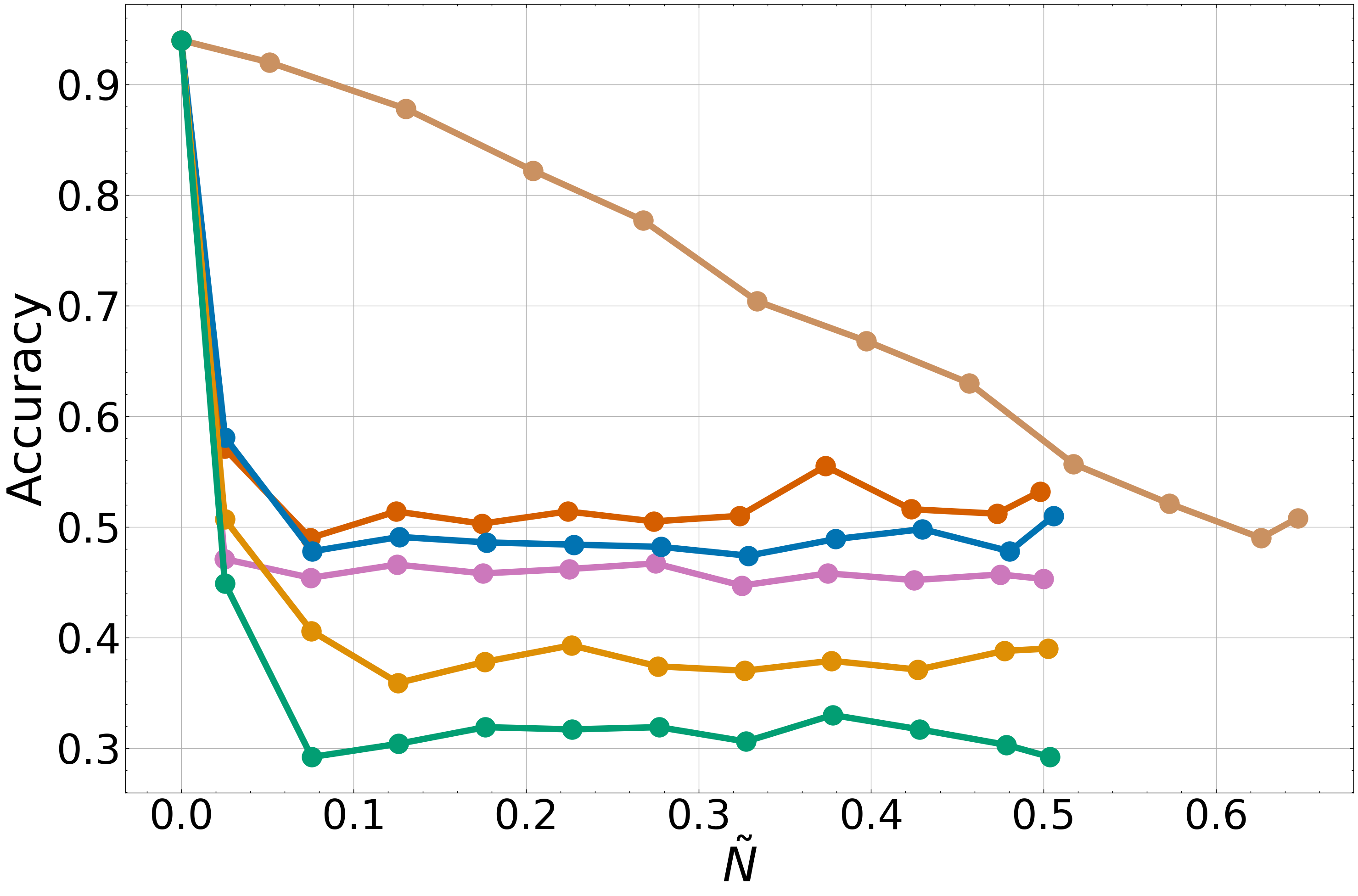}
         \vspace{-0.5cm}
         \caption{Bi-LSTM }
     \end{subfigure}
     \hfill
     \begin{subfigure}[b]{0.49\textwidth}
         \centering
         \includegraphics[width=\textwidth]{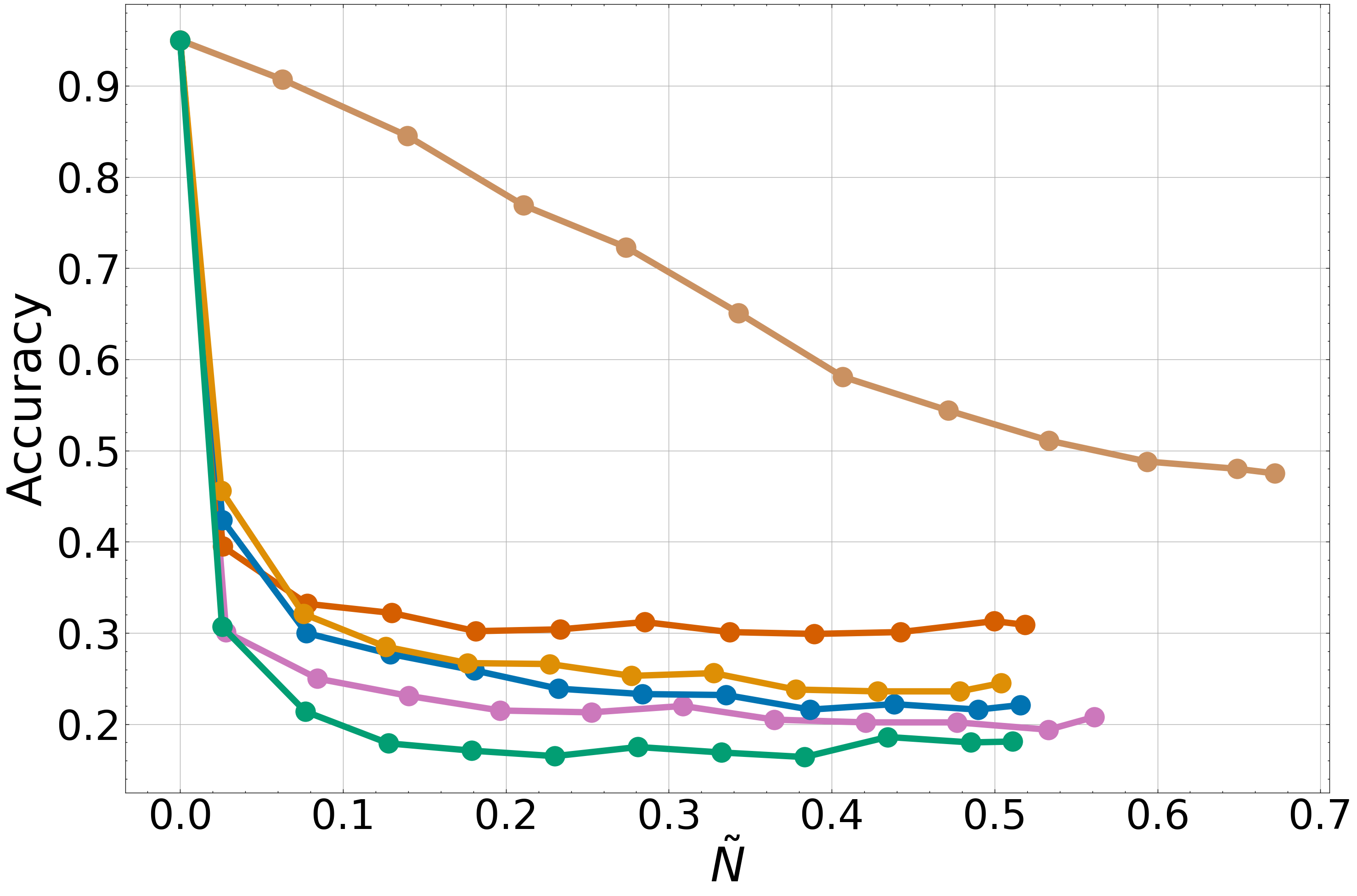}
         \vspace{-0.5cm}
         \caption{CNN}
     \end{subfigure}
     \hfill
     \begin{subfigure}[b]{0.49\textwidth}
         \centering
         \includegraphics[width=\textwidth]{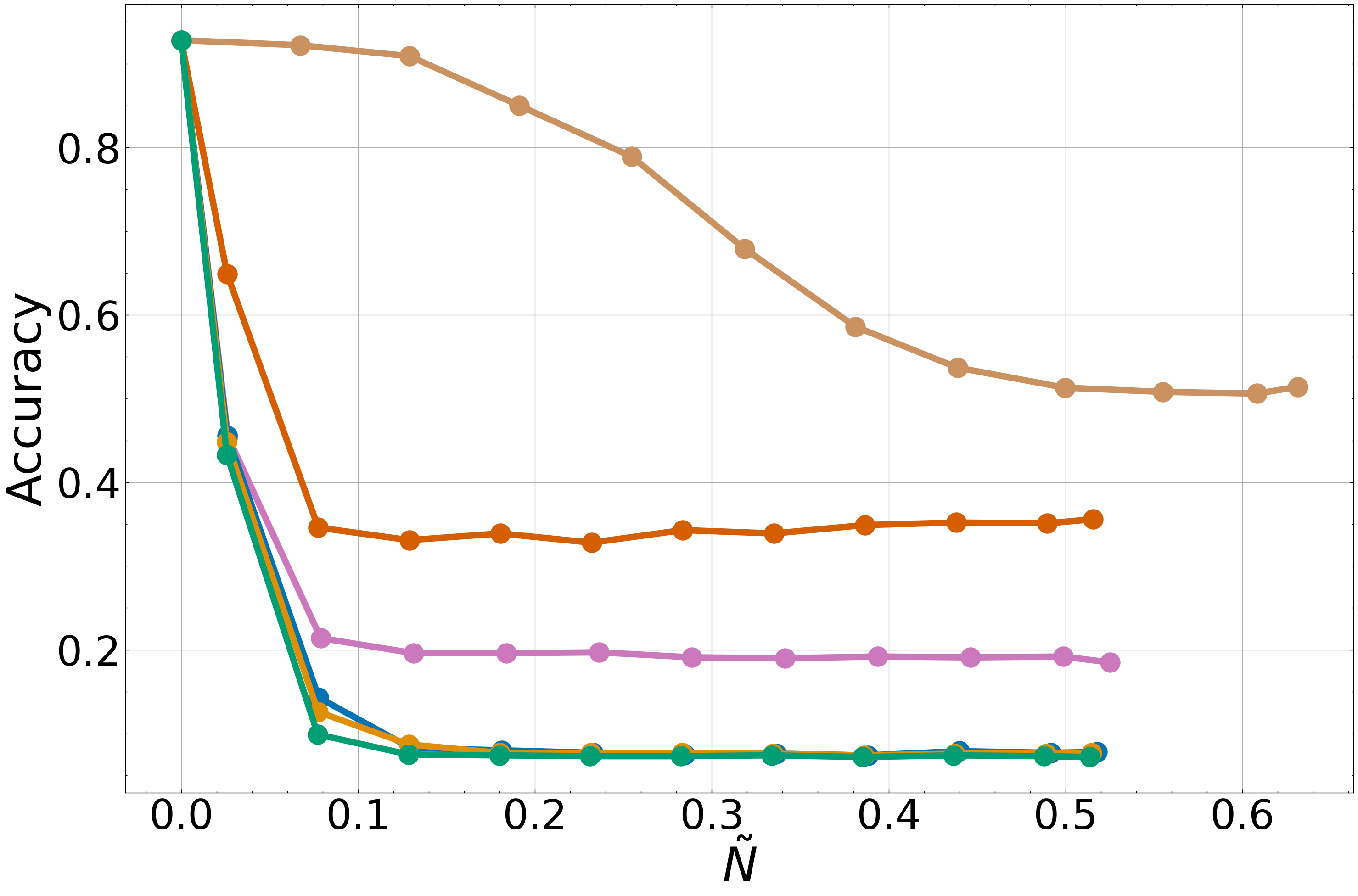}
         \vspace{-0.5cm}
         \caption{Transformer}
     \end{subfigure}
     \hfill
      \begin{subfigure}[b]{0.4\textwidth}
         \centering
         \includegraphics[width=\textwidth]{figures/main/legend.png}
         \caption*{}
     \end{subfigure}
     \caption{\textbf{Change in accuracy as a function of the ratio of points removed with respect to the total number of time steps in the sample, $\tilde{N}$ for the six interpretability methods considered using the synthetic dataset.} Results depicted for (a) Bi-LSTM, (b) CNN and (C) Transformer.}
        \label{fig:accuracy_synthetic}
     \end{figure}
\begin{figure}[H]
     \centering
     \begin{subfigure}[b]{0.49\textwidth}
         \centering
         \includegraphics[width=\textwidth]{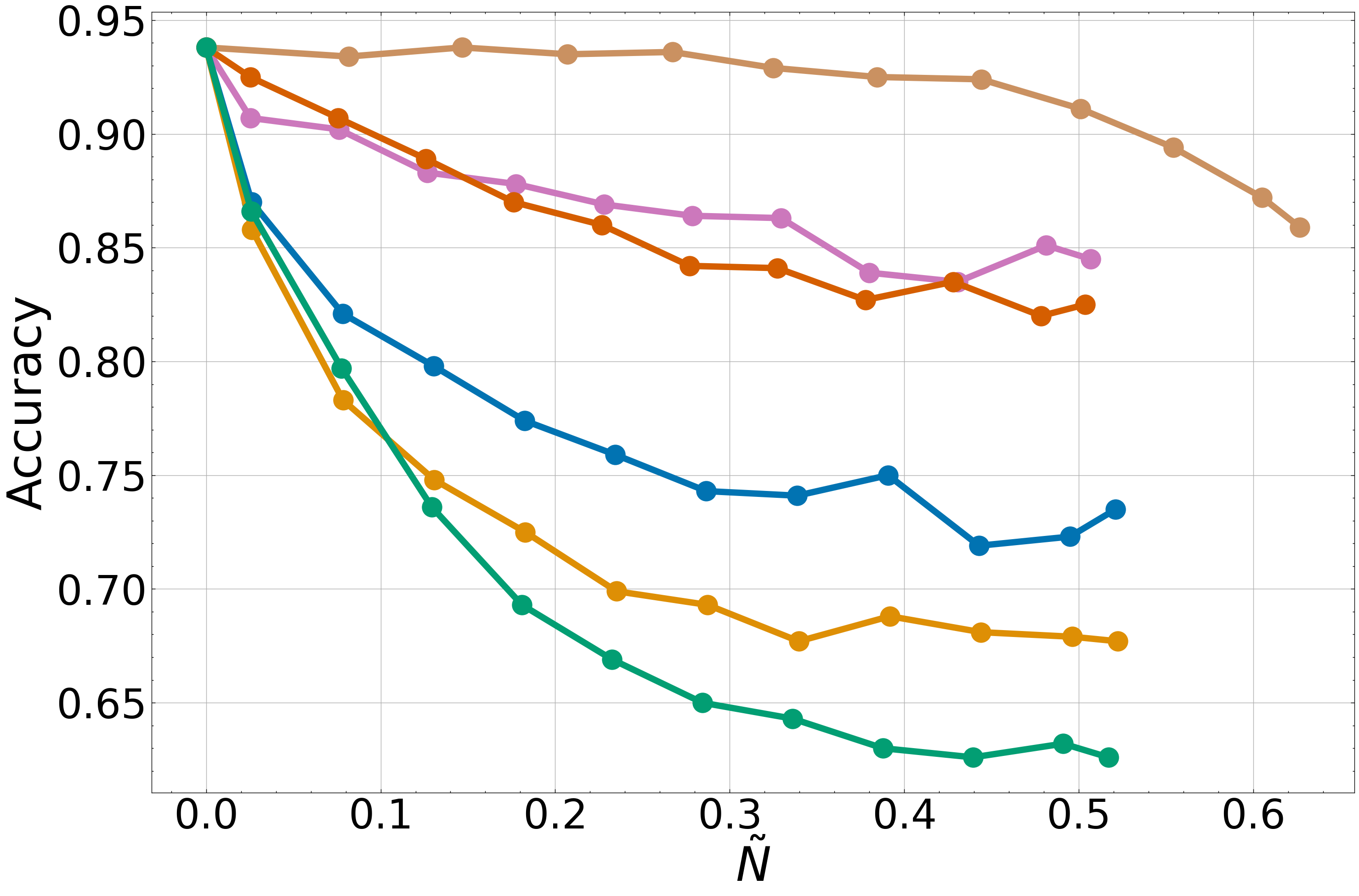}
         \vspace{-0.5cm}
         \caption{Bi-LSTM }
     \end{subfigure}
     \hfill
     \begin{subfigure}[b]{0.49\textwidth}
         \centering
         \includegraphics[width=\textwidth]{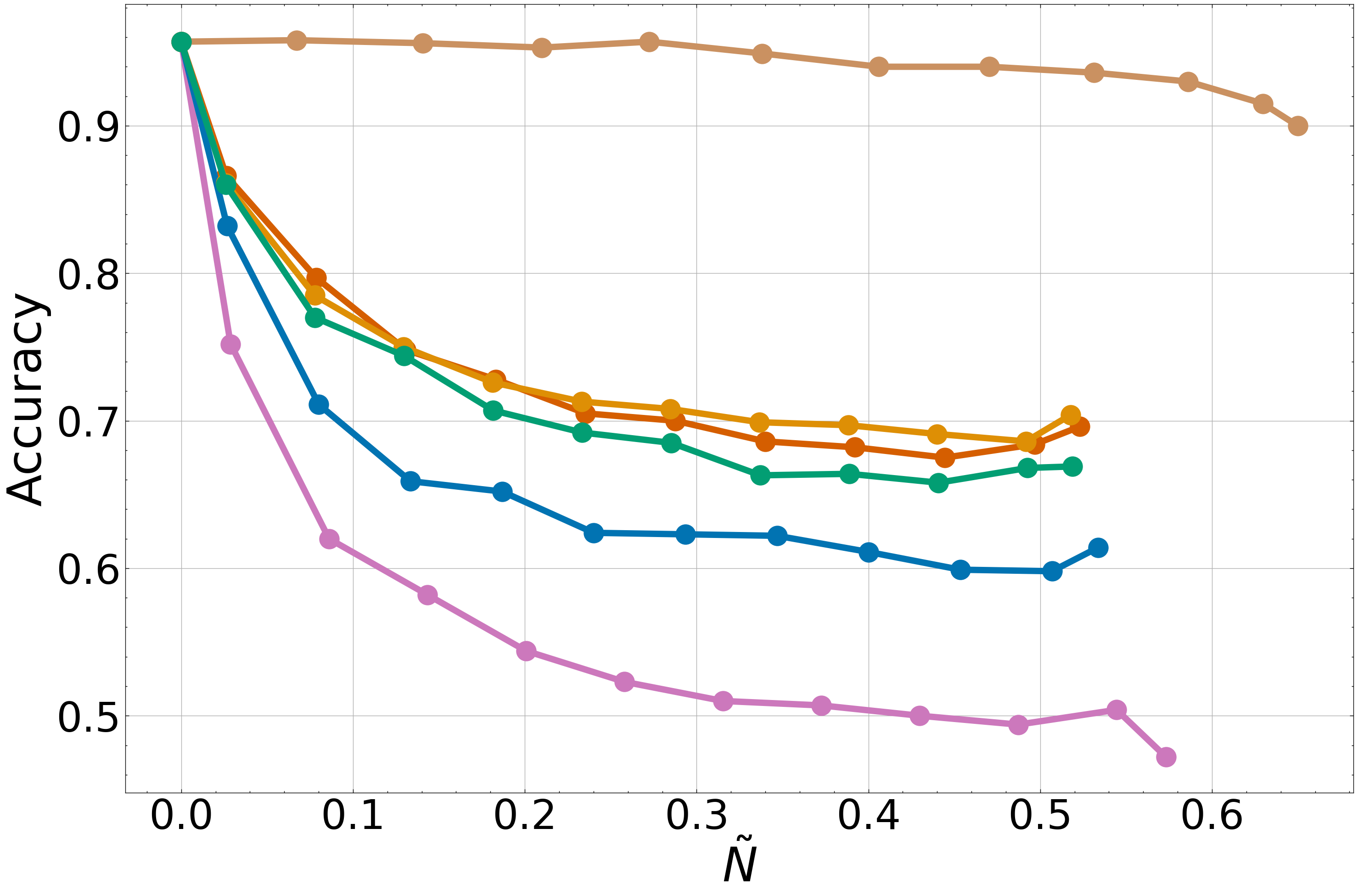}
         \vspace{-0.5cm}
         \caption{CNN}
     \end{subfigure}
     \hfill
     \begin{subfigure}[b]{0.49\textwidth}
         \centering
         \includegraphics[width=\textwidth]{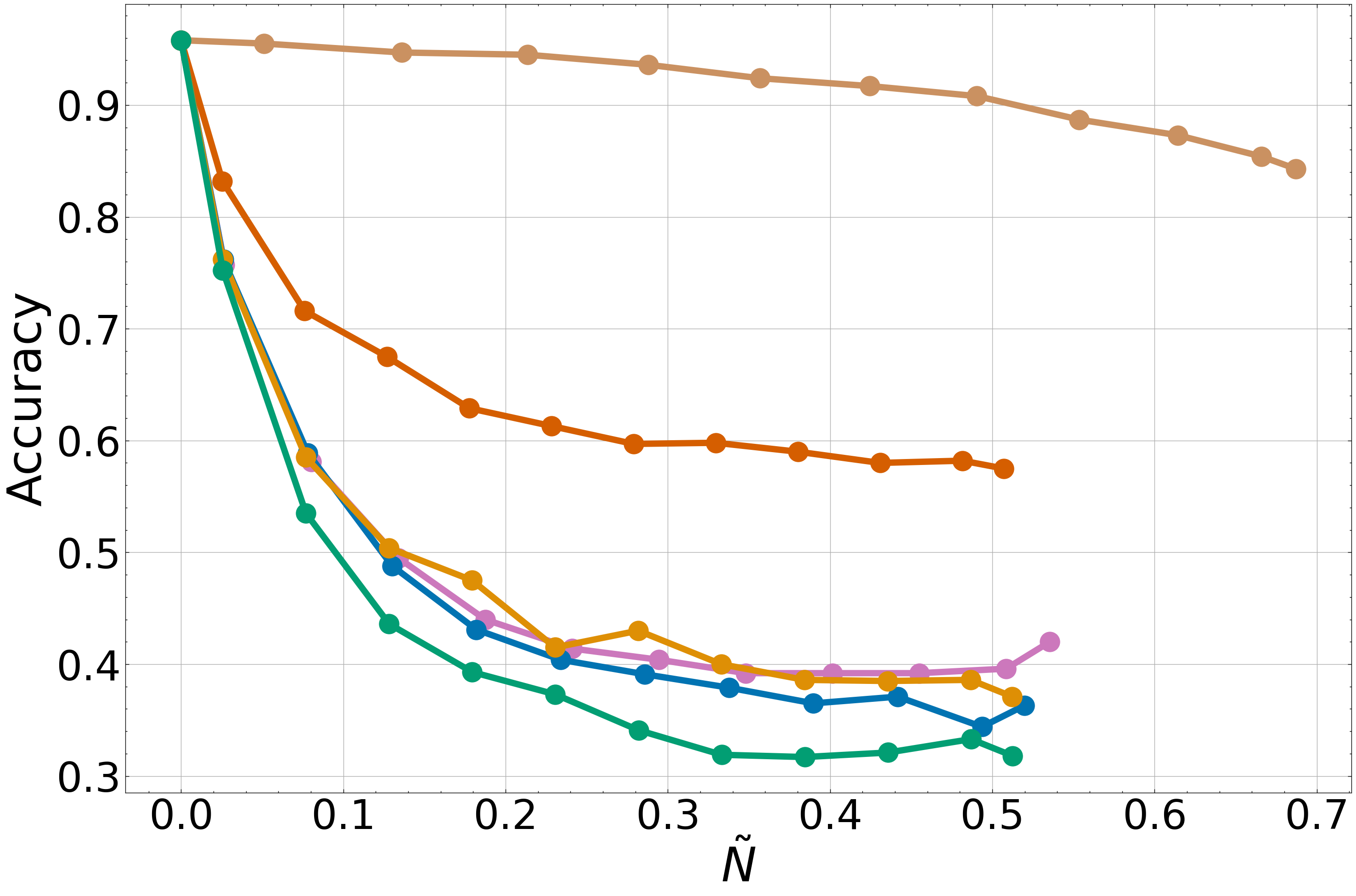}
         \vspace{-0.5cm}
         \caption{Transformer}
     \end{subfigure}
     \hfill
      \begin{subfigure}[b]{0.4\textwidth}
         \centering
         \includegraphics[width=\textwidth]{figures/main/legend.png}
         \caption*{}
     \end{subfigure}
     \caption{\textbf{Change in accuracy as a function of the ratio of points removed with respect to the total number of time steps in the sample, $\tilde{N}$ for the six interpretability methods considered using the ECG dataset.} Results depicted for (a) Bi-LSTM, (b) CNN and (C) Transformer.}
        \label{fig:accuracy_ecg}
     \end{figure}
\begin{figure}[H]
     \centering
     \begin{subfigure}[b]{0.49\textwidth}
         \centering
         \includegraphics[width=\textwidth]{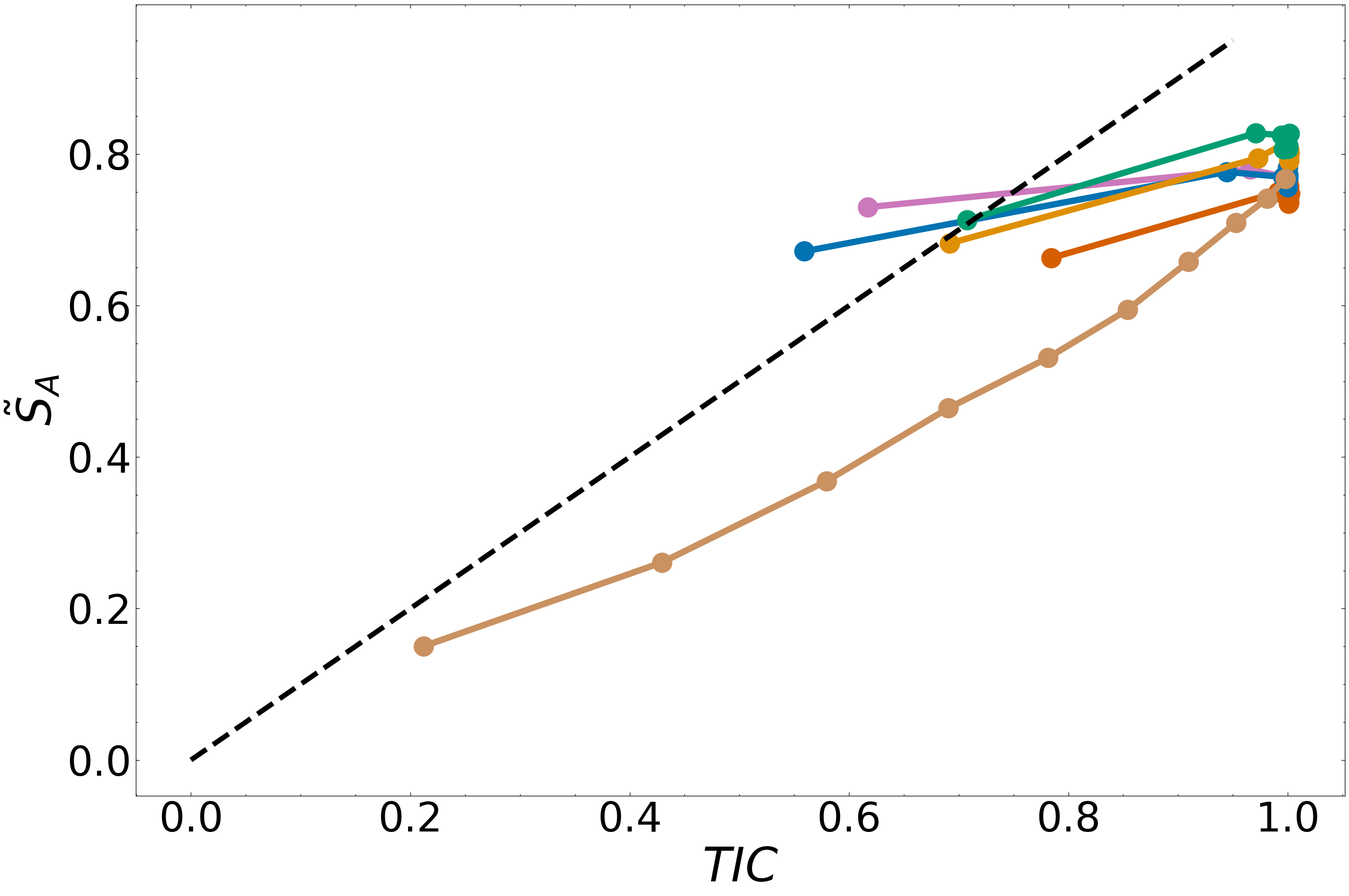}
         \vspace{-0.5cm}
         \caption{Bi-LSTM }
     \end{subfigure}
     \hfill
     \begin{subfigure}[b]{0.49\textwidth}
         \centering
         \includegraphics[width=\textwidth]{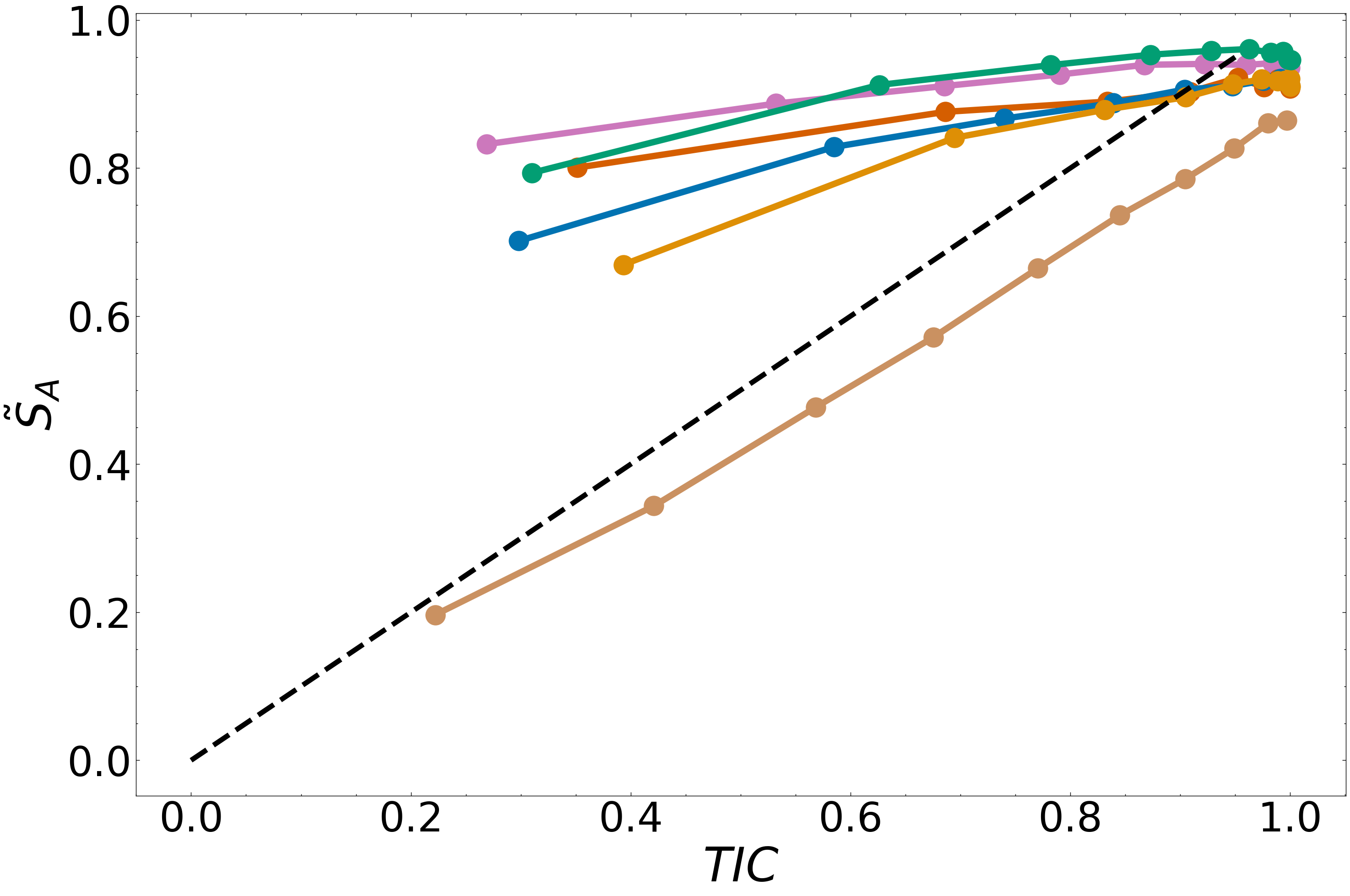}
         \vspace{-0.5cm}
         \caption{CNN}
     \end{subfigure}
     \hfill
     \begin{subfigure}[b]{0.49\textwidth}
         \centering
         \includegraphics[width=\textwidth]{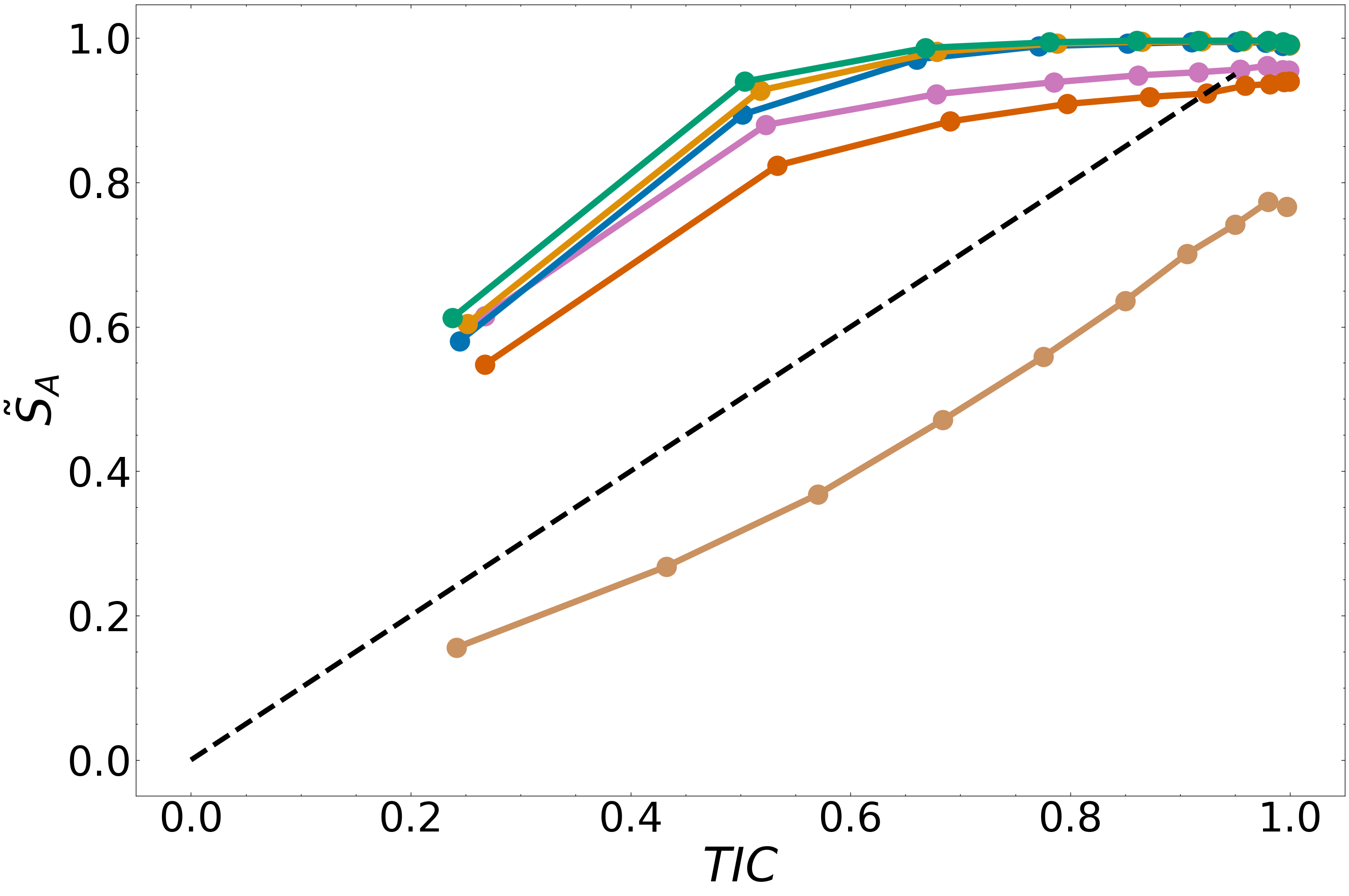}
         \vspace{-0.5cm}
         \caption{Transformer}
     \end{subfigure}
     \hfill
      \begin{subfigure}[b]{0.4\textwidth}
         \centering
         \includegraphics[width=\textwidth]{figures/main/legend.png}
         \caption*{}
     \end{subfigure}
     \caption{\textbf{$\tilde{S}_A$ as a function of the TIC index for the six interpretability methods considered using the synthetic dataset. }Results depicted for (a) Bi-LSTM, (b) CNN and (c) Transformer.}
        \label{fig:rel-attribution_synthetic}
     \end{figure}
\begin{figure}[H]
    \centering
    \includegraphics[width=0.9\textwidth]{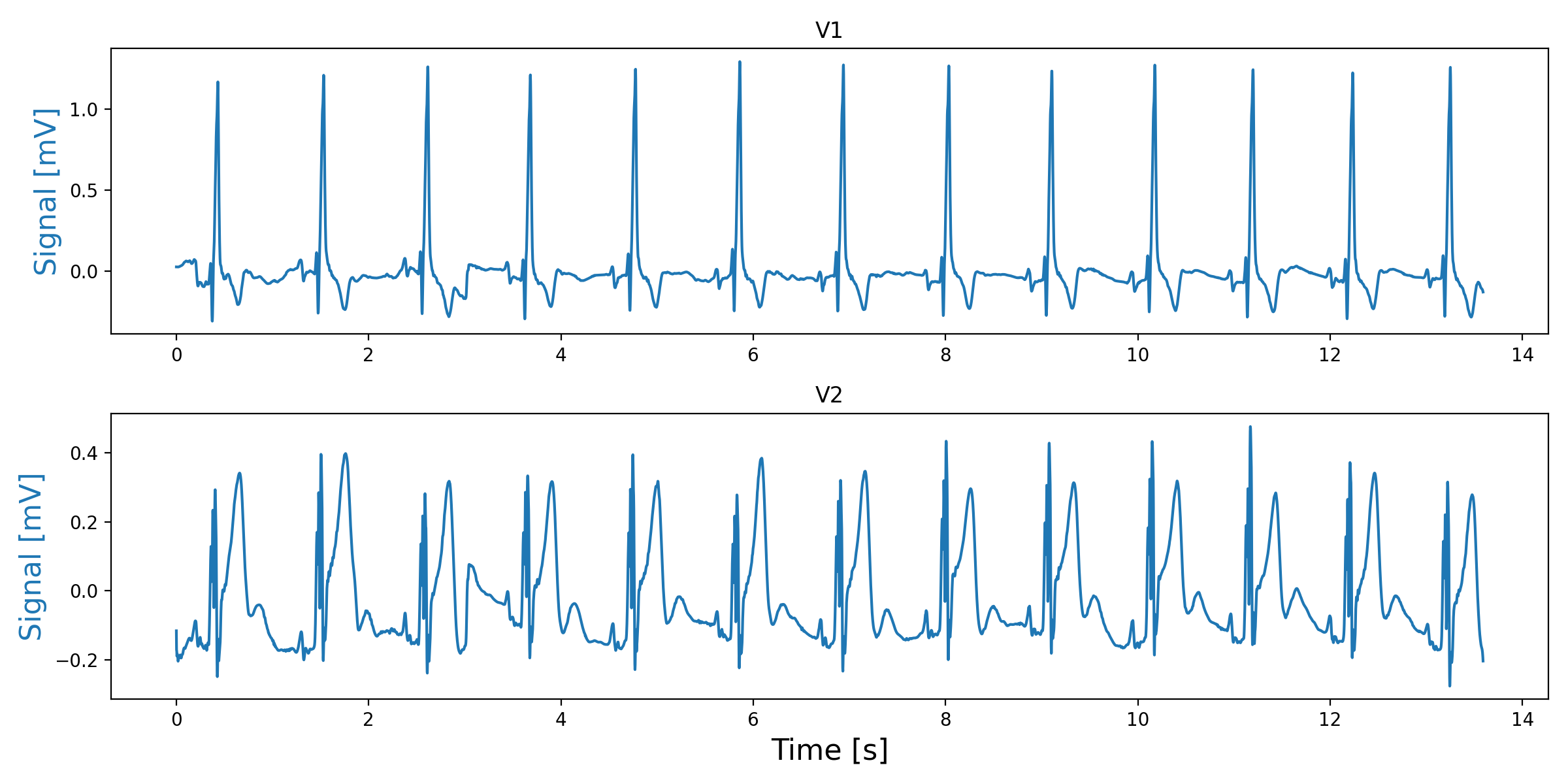}
    \caption{\textbf{Raw ECG signal for two selected lead from a given sample}}
    \label{fig:unprocessed_ECG}
\end{figure}

\begin{figure}[H]
    \centering
    \includegraphics[width=0.9\textwidth]{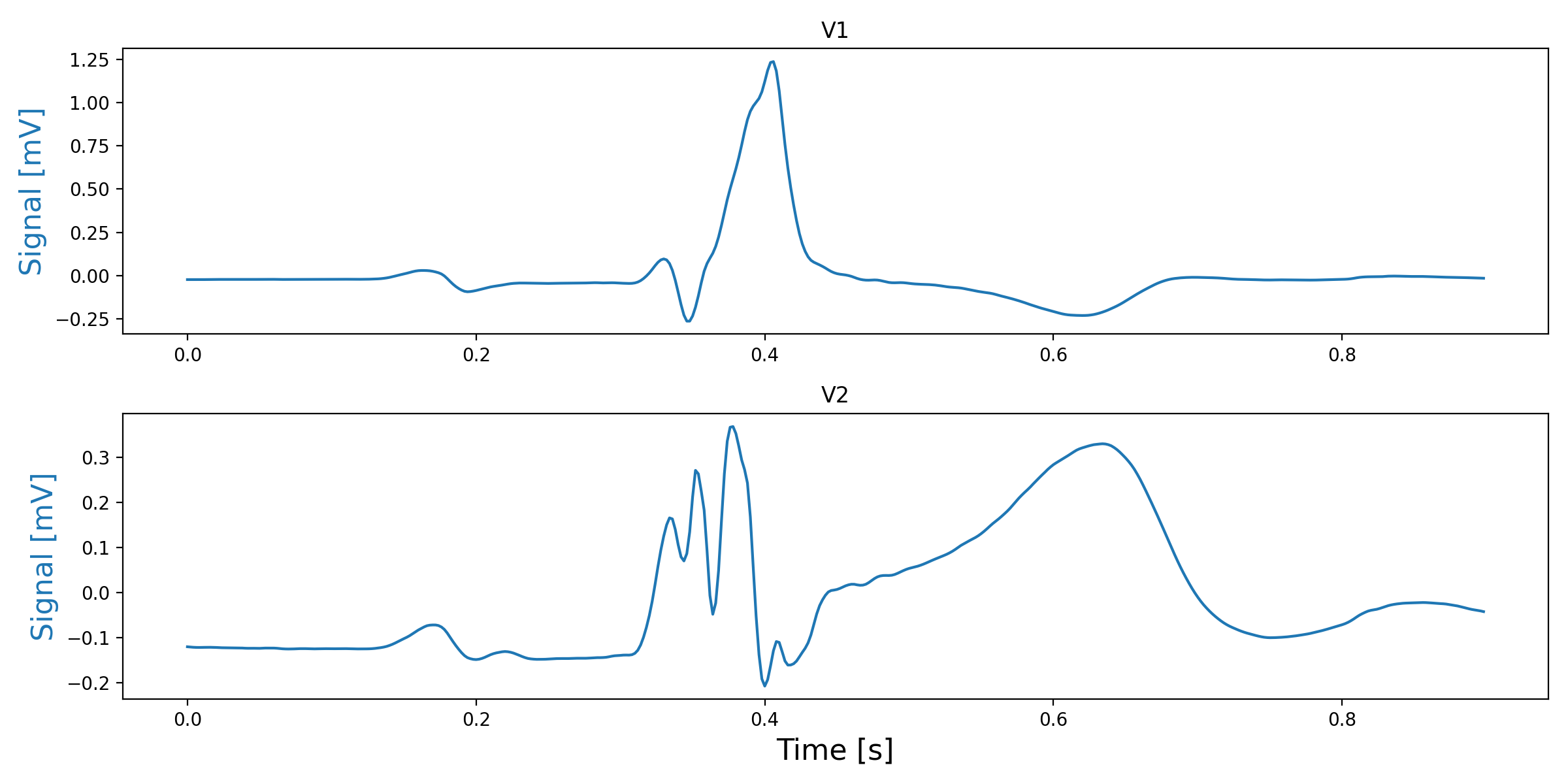}
    \caption{\textbf{Processed ECG signal for two selected lead from a given sample}}
    \label{fig:processed_ECG}
\end{figure}

\newpage

\renewcommand{\figurename}{Figure}
\renewcommand{\tablename}{Table}

\setcounter{section}{0}
\setcounter{figure}{0}
\renewcommand{\figurename}{Supplementary information figure}

\setcounter{table}{0}
\renewcommand{\tablename}{Supplementary information table}
\renewcommand*{\theHsection}{chX.\the\value{section}}
\section*{Supplementary Information}
\section{Results for the FordA dataset}
\subsection{$\tilde{S} - \tilde{N}$ curves}

\begin{figure}[H]
    \centering
    \includegraphics[width=\textwidth]{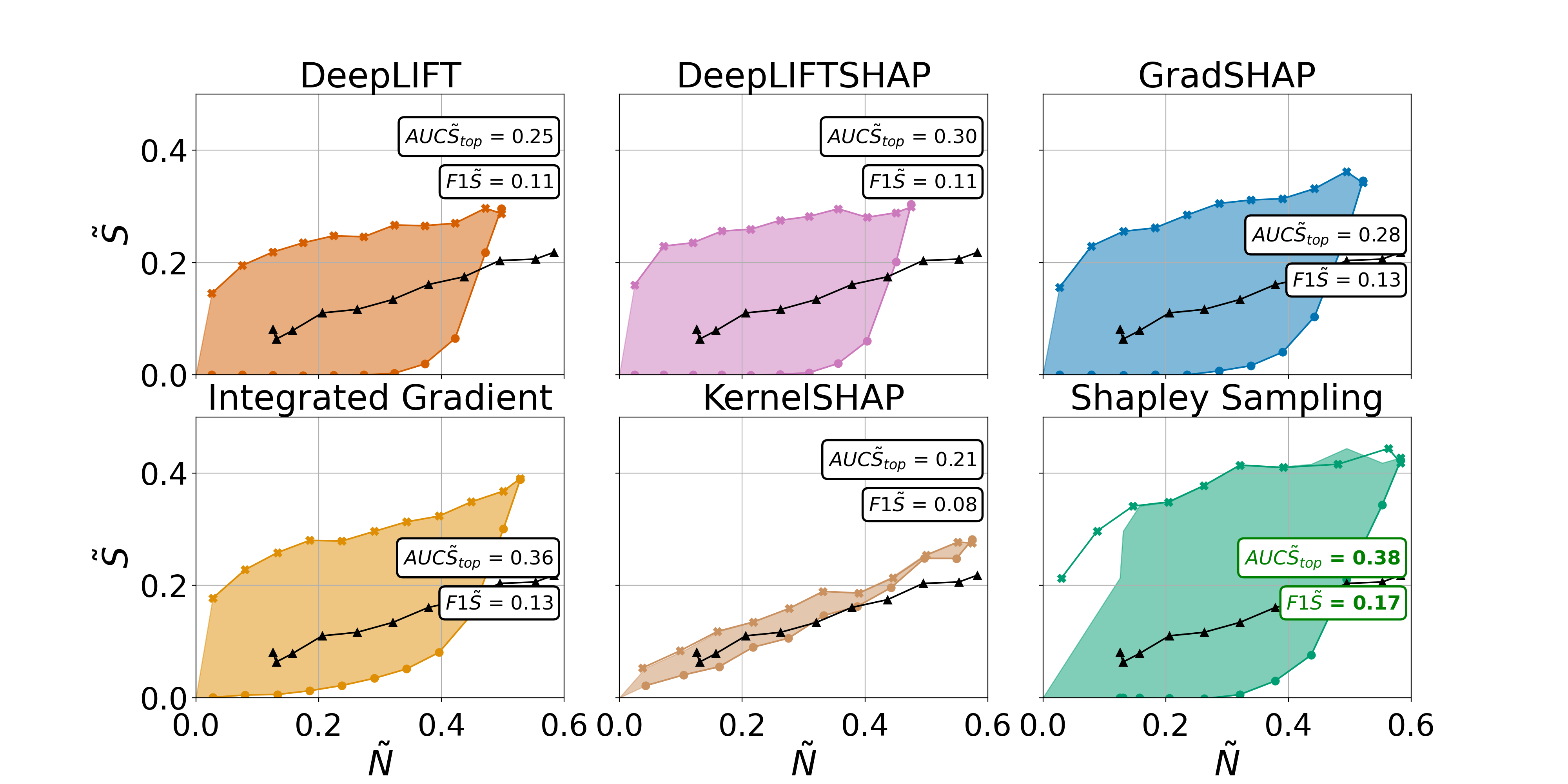 }
    \caption{\textbf{$\tilde{S}$ as a function of the ratio of points removed with respect to the total number of time steps in the sample, $\tilde{N}$.} Each subfigure represents one of the six interpretability methods considered for a Bi-LSTM trained on the Ford A dataset.}
    \label{fig:forda-dataset-identification_bilstm}
\end{figure}
\begin{figure}[H]
     \centering
     \includegraphics[width=\textwidth]{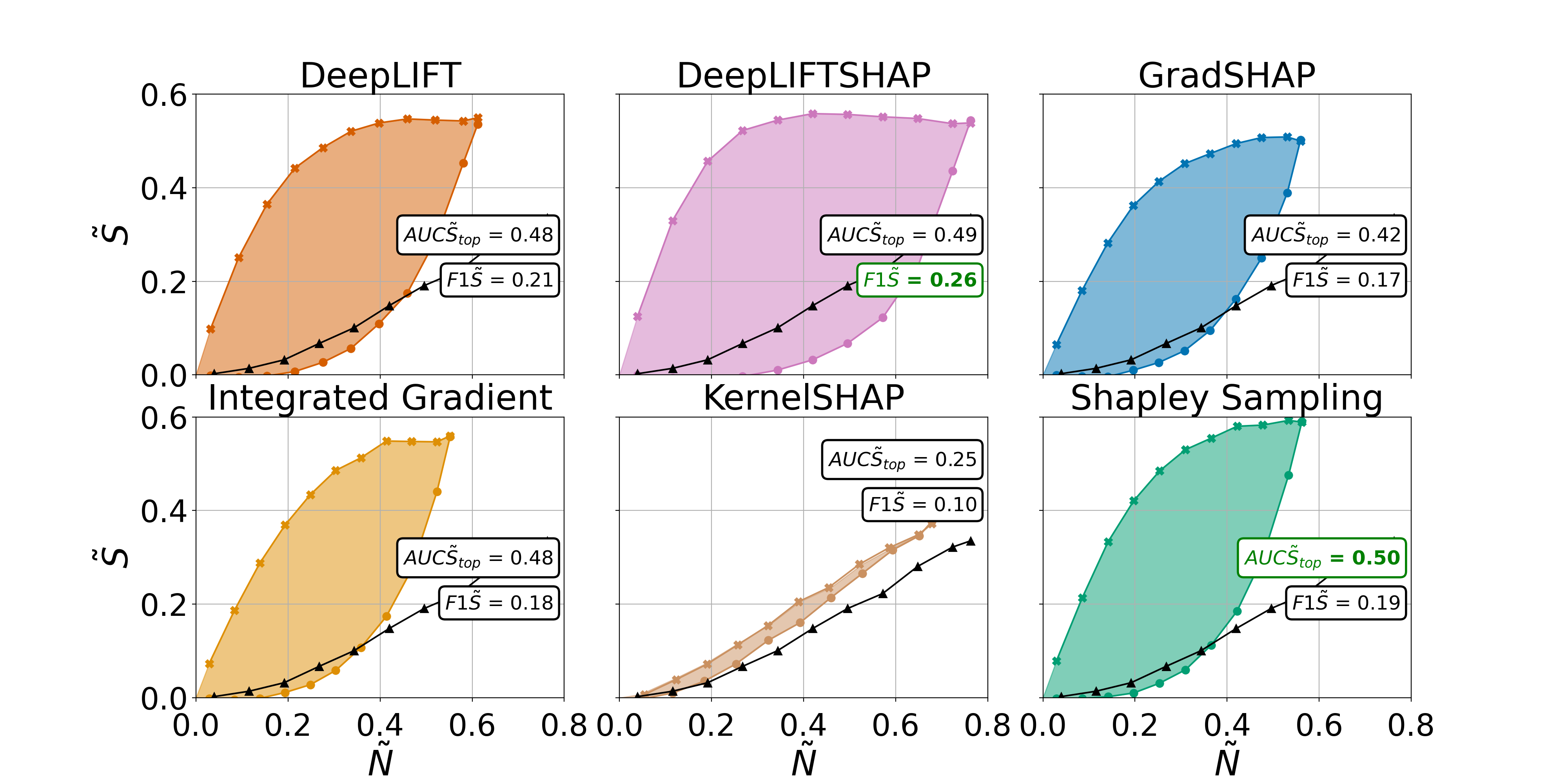}
    \caption{\textbf{$\tilde{S}$ as a function of the ratio of points removed with respect to the total number of time steps in the sample, $\tilde{N}$.} Each subfigure represents one of the six interpretability methods considered for a CNN trained on the Ford A dataset.}
    \label{fig:forda-dataset-identification_cnn}
\end{figure}
\begin{figure}[H]
     \centering
     \includegraphics[width=\textwidth]{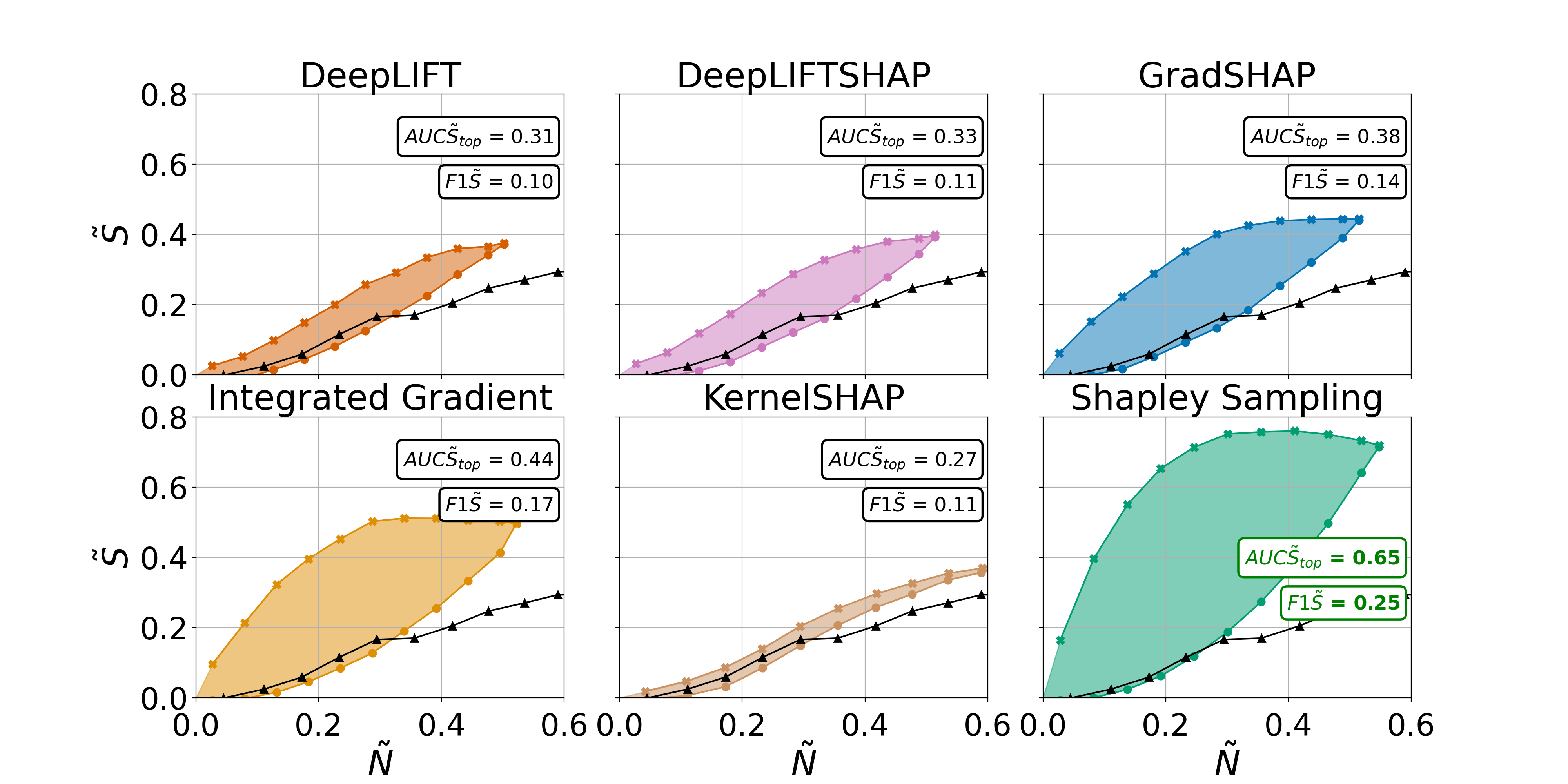}
    \caption{\textbf{$\tilde{S}$ as a function of the ratio of points removed with respect to the total number of time steps in the sample, $\tilde{N}$.} Each subfigure represents one of the six interpretability methods considered for a Transformer trained on the Ford A dataset.}
    \label{fig:forda-dataset-identification_transformer}
\end{figure}

\subsection{Accuracy drop}

\begin{figure}[H]
     \centering
     \begin{subfigure}[b]{0.49\textwidth}
         \centering
         \includegraphics[width=\textwidth]{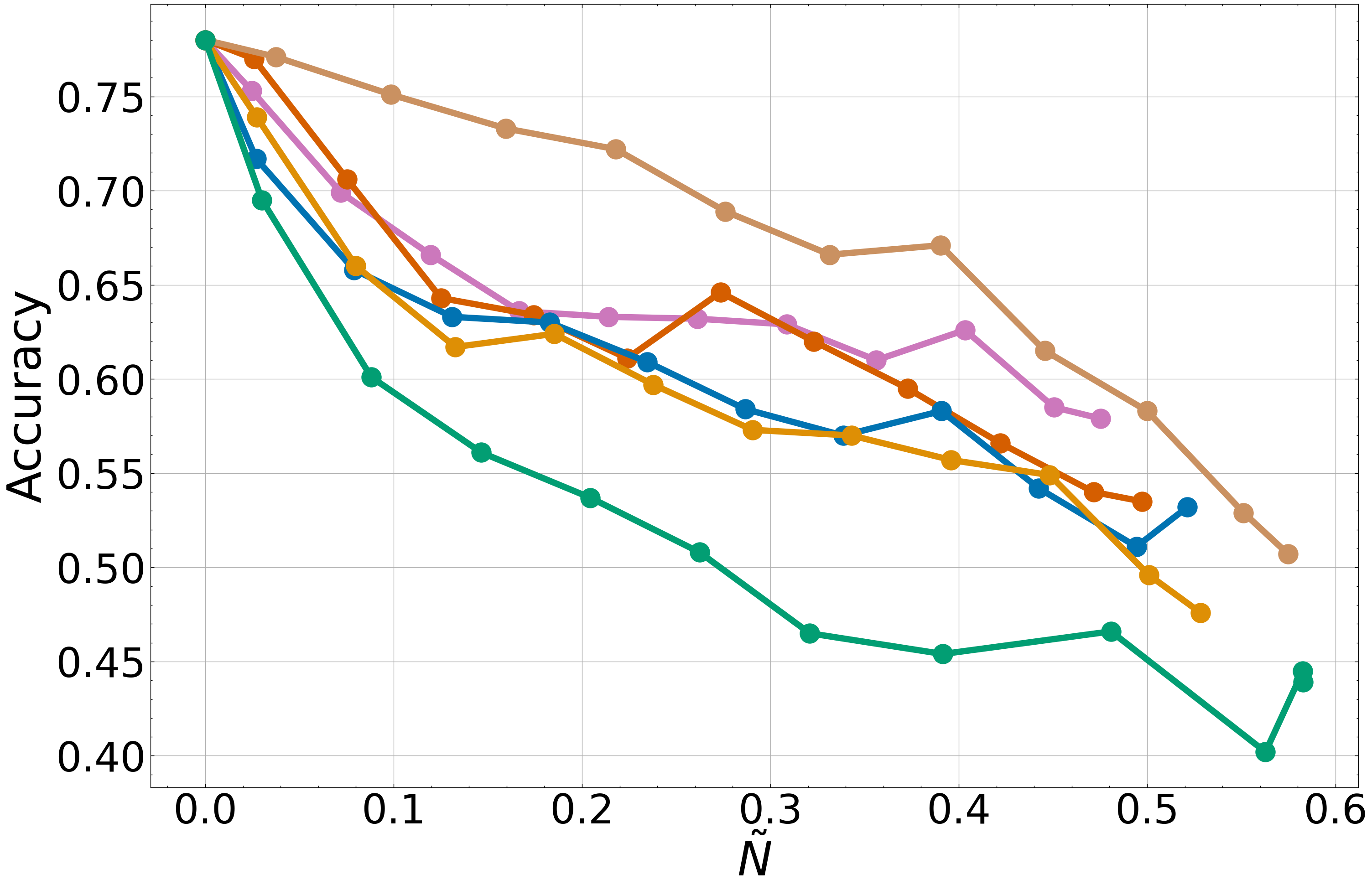}
         \vspace{-0.5cm}
         \caption{Bi-LSTM }
     \end{subfigure}
     \hfill
     \begin{subfigure}[b]{0.49\textwidth}
         \centering
         \includegraphics[width=\textwidth]{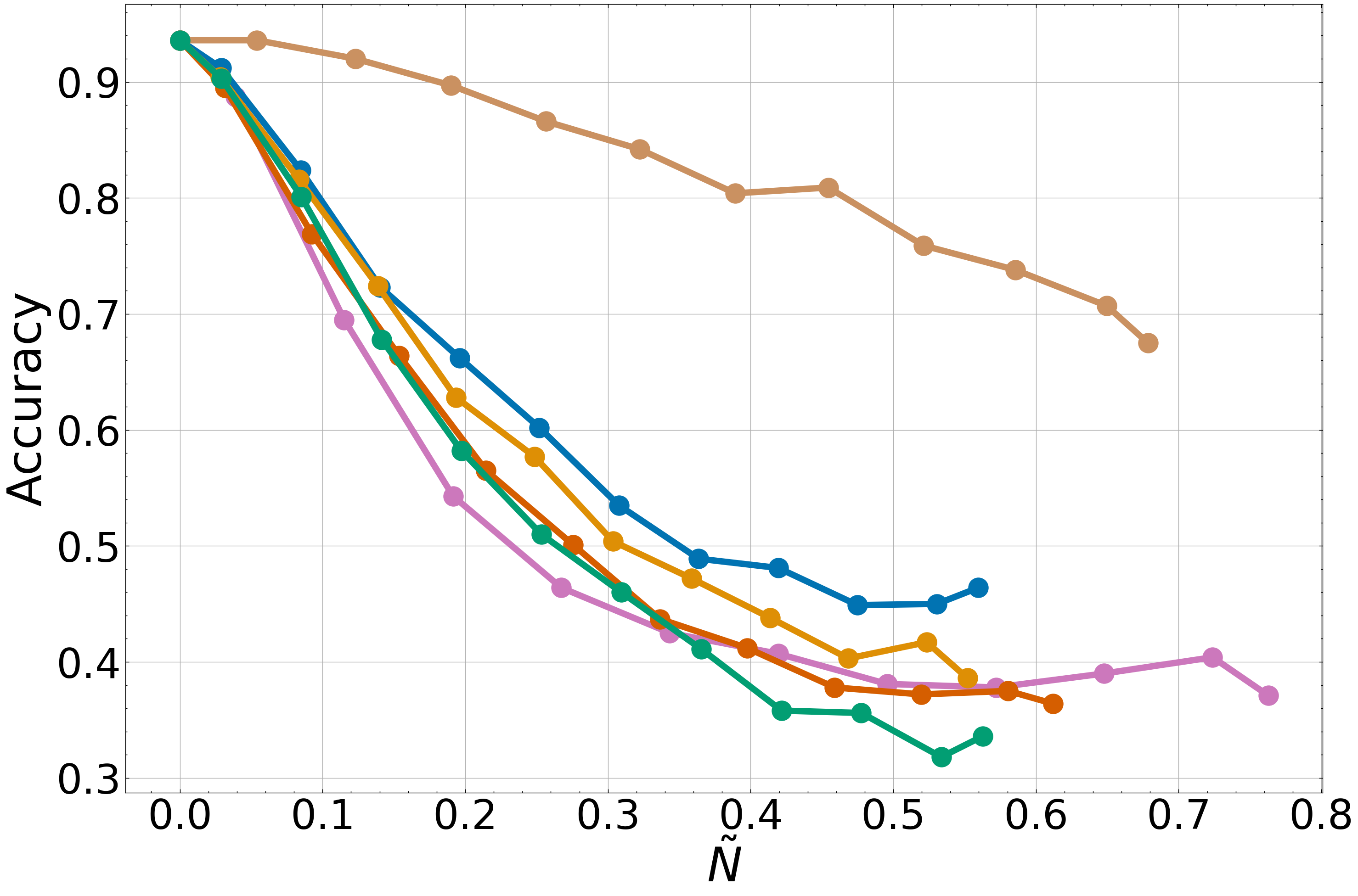}
         \vspace{-0.5cm}
         \caption{CNN}
     \end{subfigure}
     \hfill
     \begin{subfigure}[b]{0.49\textwidth}
         \centering
         \includegraphics[width=\textwidth]{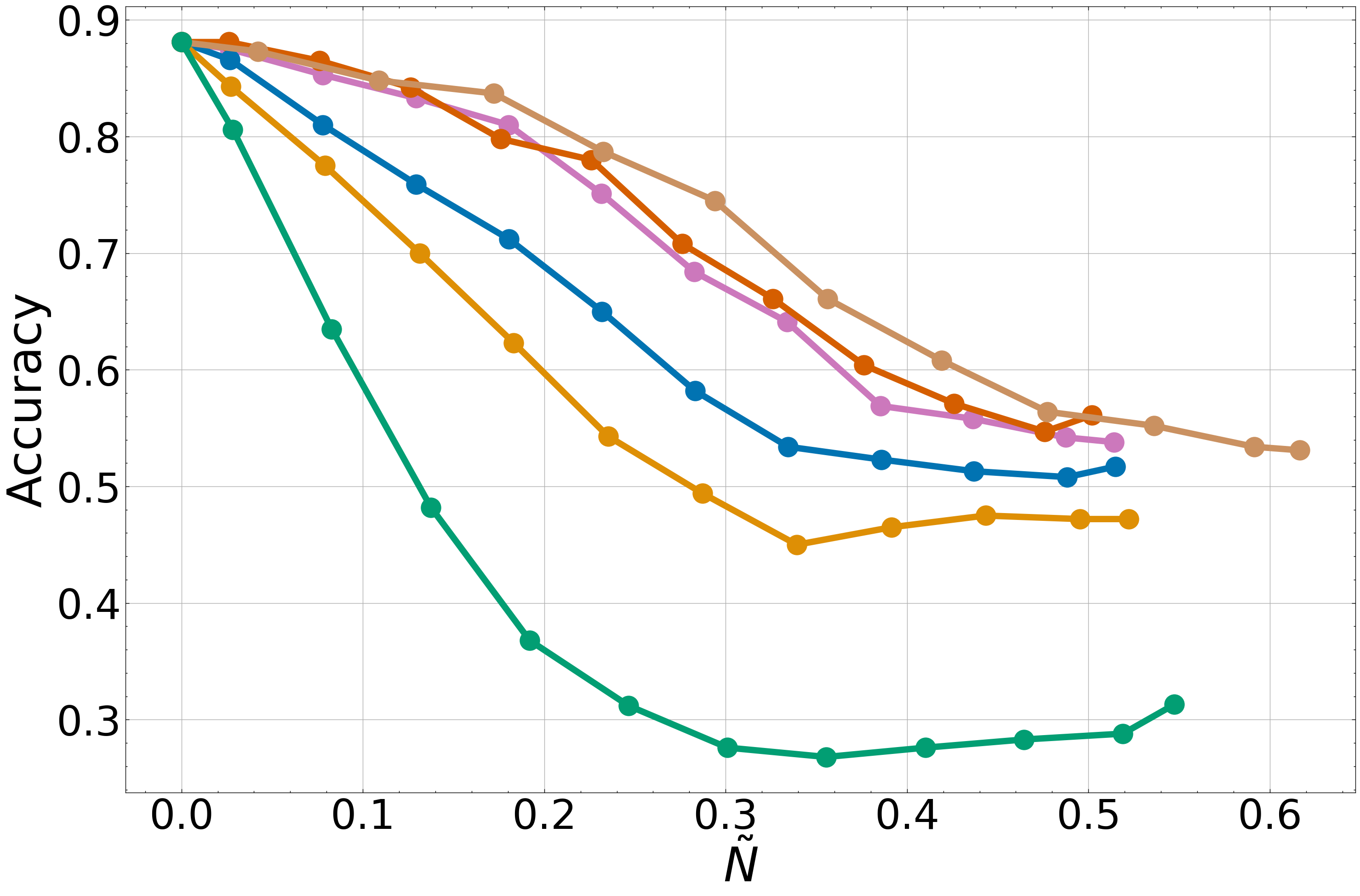}
         \vspace{-0.5cm}
         \caption{Transformer}
     \end{subfigure}
     \hfill
      \begin{subfigure}[b]{0.4\textwidth}
         \centering
         \includegraphics[width=\textwidth]{figures/main/legend.png}
         \caption*{}
     \end{subfigure}
     \caption{Change in accuracy as a function of the ratio of points removed with respect to the total number of time steps in the sample, $\tilde{N}$ for the six interpretability methods considered using the Ford A dataset and for (a) Bi-LSTM, (b) CNN and (c) Transformer.}
        \label{fig:accuracy_forda}
     \end{figure} 
     
\subsection{$\tilde{S}_A$ vs. TIC curves}
\begin{figure}[H]
     \centering
     \begin{subfigure}[b]{0.49\textwidth}
         \centering
         \includegraphics[width=\textwidth]{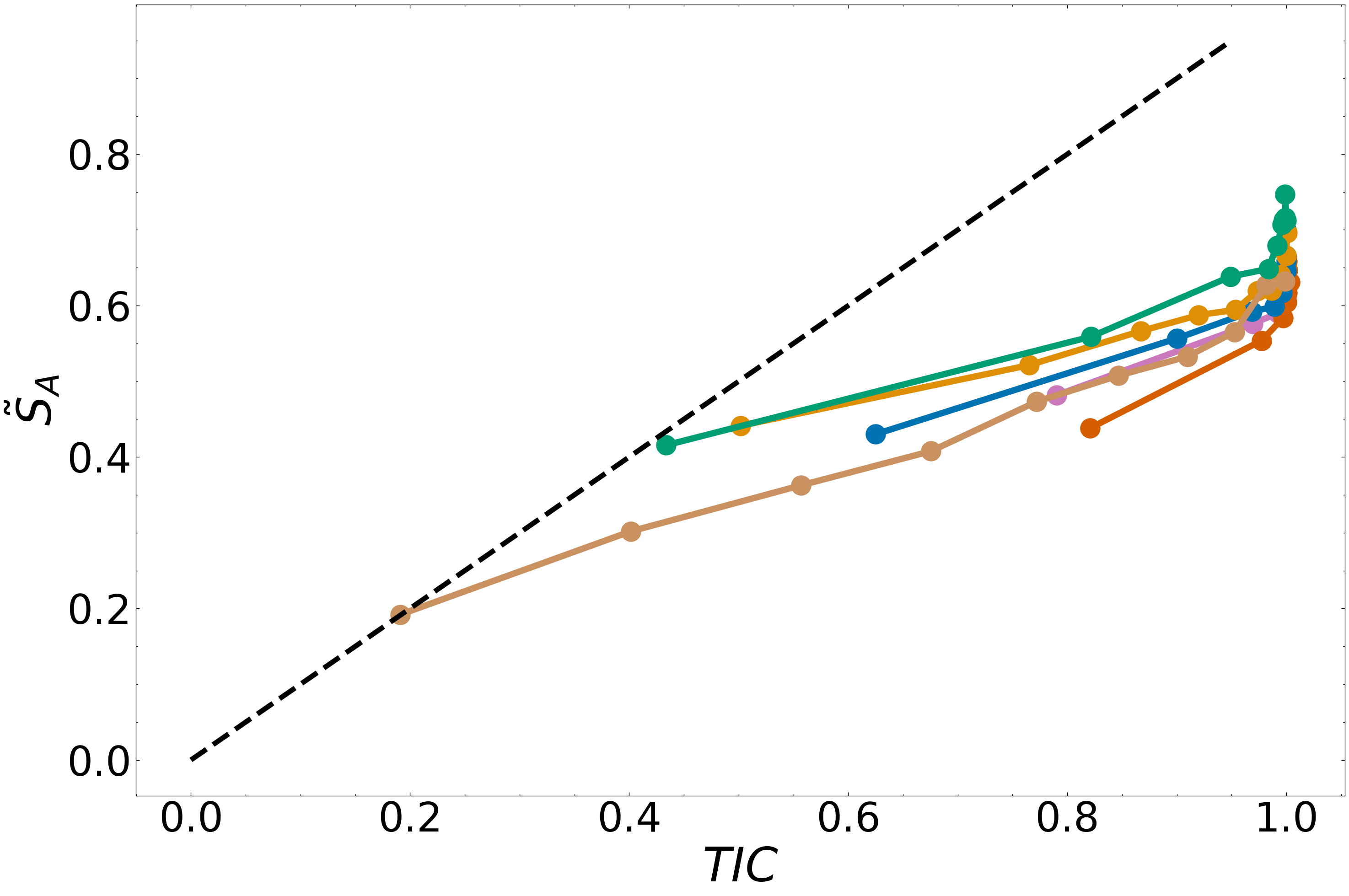}
         \vspace{-0.5cm}
         \caption{Bi-LSTM }
     \end{subfigure}
     \hfill
     \begin{subfigure}[b]{0.49\textwidth}
         \centering
         \includegraphics[width=\textwidth]{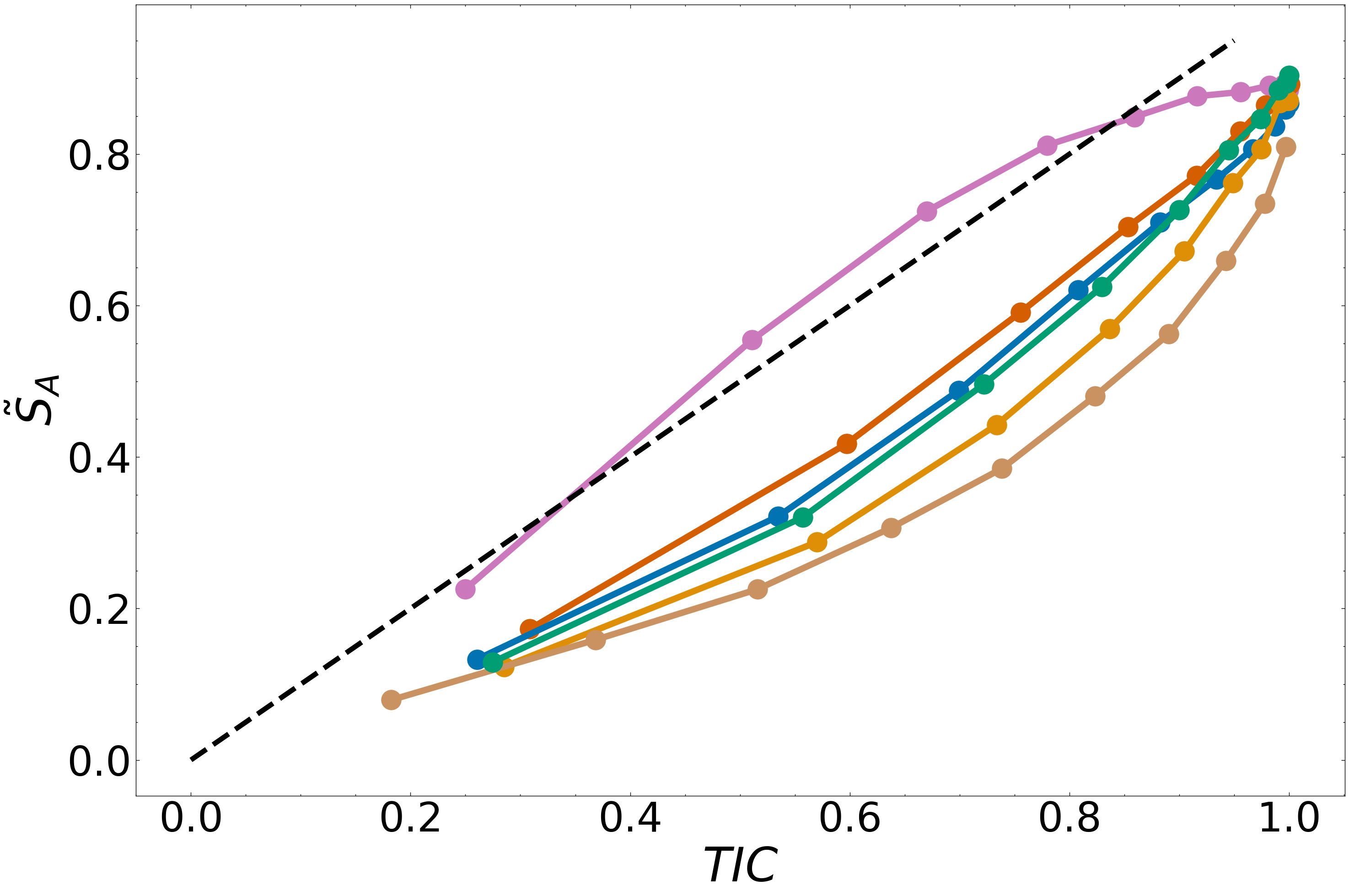}
         \vspace{-0.5cm}
         \caption{CNN}
     \end{subfigure}
     \hfill
     \begin{subfigure}[b]{0.49\textwidth}
         \centering
         \includegraphics[width=\textwidth]{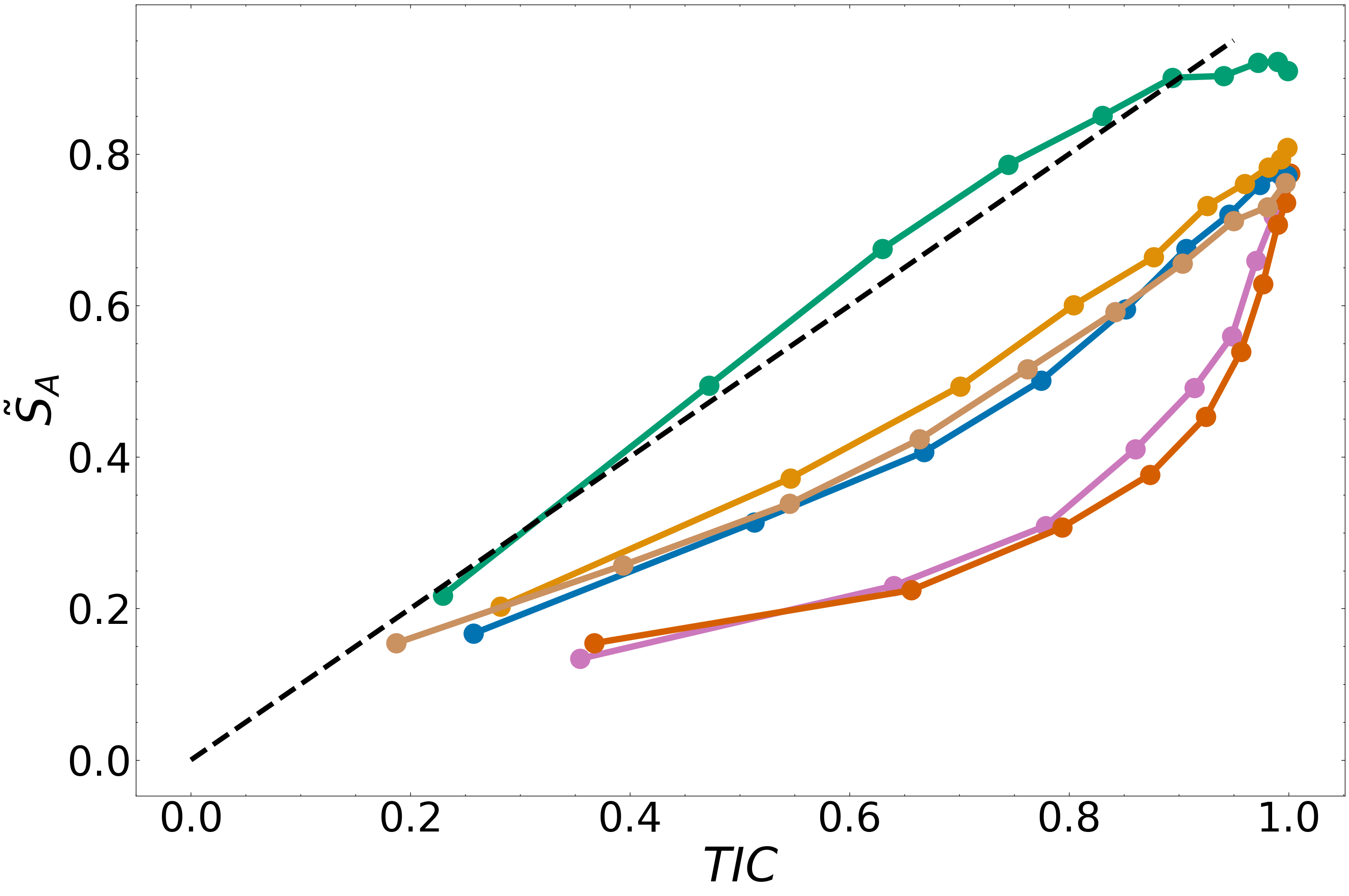}
         \vspace{-0.5cm}
         \caption{Transformer}
     \end{subfigure}
     \hfill
      \begin{subfigure}[b]{0.4\textwidth}
         \centering
         \includegraphics[width=\textwidth]{figures/main/legend.png}
         \caption*{}
     \end{subfigure}
     \caption{$\tilde{S}_A$ as a function of the TIC index for the six interpretability methods considered using the Ford A dataset and for (a) Bi-LSTM, (b) CNN and (C) Transformer.}
        \label{fig:rel-attribution_forda}
     \end{figure}

\section{Models' configuration}
\label{sec:appendix_model_configuration}
Three models per dataset were trained: CNN, Bi-LSTM, and transformer.  The transformer architecture follows the adaptation for time series classification presented by Zerveas et al.~\cite{zerveas2020transformer}. All models were trained for a maximum of 200 epochs with an early stop criterion monitoring the loss on the validation set. In addition, all models were trained with the RAdam optimiser \cite{radam}. An initial configuration was tested for each architecture on each of the three datasets. If the classification accuracy measured on the test set was below 90\%, an hyperparameter search was performed to find the optimal architecture. The domain for the hyperparameter search is presented in table~\ref{table:hyperparameter_space}. The hyperparameter search was performed using the \textit{Ray Tune} library using Bayesian optimisation with 64 samples and HyperBand scheduler.

\begin{table}[H]
\centering
\caption{Hyperparmeter search domain for the models trained on the ECG datasets}
\label{table:hyperparameter_space}
\begin{tabular}{@{}llll@{}}
\toprule
            & Parameter        & Search Space           & Sampling   \\ \midrule
Common      & learning rate    & {[}1e-5,1e-2{]}        & Loguniform \\
            & batch size       & {[}64, 128, 256{]}     & Choice     \\ \hline
CNN         & nb. layers       & {[}3,4{]}              & Choice     \\
            & filters          & {[}32, 64, 128, 256{]} & Choice     \\
            & kernel size      & {[}3,5,7,11,15{]}      & Choice     \\\hline
Bi-LSTM     & nb. layers       & {[}3,4{]}              & Choice     \\
            & filters          & {[}32, 64, 128 {]}     & Choice     \\ \hline
Transformer & input dropout    & {[}0,0.5{]}            & Uniform    \\
            & nb encoder layer & {[}2,3,4,5{]}          & Choice     \\
            & nb head          & {[}2,4,6,8{]}          & Choice     \\
            & Projection size  & {[}64,128{]}           & Choice     \\
            & MLP dim.          & {[}64,128,256{]}      & Choice     \\ \bottomrule
\end{tabular}
\end{table}

\subsection{Models trained on the synthetic dataset}
The configurations of the models trained on the synthetic dataset are presented in tables \ref{table:bilstm_parameters_synthetic}, \ref{table:cnn_parameters_synthetic}, \ref{table:transformer_parameters_synthetic}  while the the batch size and learning rate used are presented in table~\ref{table:synthetic_parameters}.
\begin{table}[H]
\centering
\caption{Bi-LSTM Models' hyperparameters for the synthetic datasets}
\label{table:bilstm_parameters_synthetic}
\begin{tabular}{@{}lll@{}}
\toprule
\# & Layer   & Parameters                             \\ \midrule
1  & Bi-LSTM & Units = 64   \\
2  & Dense   & Units =64, Activation = ReLU   \\
3  & BatchNorm   &    \\
4  & Bi-LSTM & Units = 128   \\
5  & Dense   & Units =128, Activation = ReLU   \\
6  & BatchNorm   &    \\
7  & Bi-LSTM & Units = 128   \\
8  & Dense   & Units =128, Activation = ReLU   \\
9  & BatchNorm   &    \\
10  & Bi-LSTM & Units = 64   \\
11 & Dense   & Units =64, Activation = ReLU   \\
12  & BatchNorm   &    \\
13  & Dense   & Units =128, Activation = ReLU   \\
14  & Dense   & Units =2    \\ \bottomrule
\end{tabular}
\end{table}

\begin{table}[H]
\centering
\caption{CNN Models' hyperparameters for the synthetic dataset}
\label{table:cnn_parameters_synthetic}
\begin{tabular}{@{}lll@{}}
\toprule
\# & Layer   & Parameters                             \\ \midrule
1  & Conv 1D & $256 \times 11$, strides =1   \\
2  & Dropout & Rate =0.3                             \\
3  & Conv 1D & $256 \times 11$, strides =1 \\
4  & Dropout & Rate =0.3                             \\
5  & Conv 1D & $256 \times 11$, strides =1 \\
7  & Average Pooling &                            \\
8  & Dense   & units =2             \\ \bottomrule
\end{tabular}
\end{table}

\begin{table}[H]
\centering
\caption{Transformer Models' hyperparameters for the synthetic datasets}
\label{table:transformer_parameters_synthetic}
\begin{tabular}{@{}cll@{}}
\toprule
\# & Layer   & Parameters                             \\ \midrule
1  & Conv 1D & $128 \times 3$, strides =1   \\
1  & Conv 1D & $64 \times 3$, strides =1   \\
3  & Positional encoding &                             \\
4  & Concatenate & Concatenate input and positional embedding  \\
5  & Transformer  Layer\footnotemark[1] &  heads =4, dim. feedforward =256, dropout =0.1, mlp dim = 128  \\
6  & Transformer  Layer\footnotemark[1] & heads  =4, dim. feedforward =256, dropout =0.1, mlp dim = 128  \\
7  & Transformer  Layer\footnotemark[1]& heads =4, dim. feedforward =256, dropout =0.1  , mlp dim = 128  \\
8  & Transformer  Layer\footnotemark[1] &heads =4, dim. feedforward =256, dropout =0.1  , mlp dim = 128  \\  
9  & Dense   & Units =128, Activation = ReLU   \\
10  & Dense   & Units =2    \\ \bottomrule
\end{tabular}
\end{table}

\begin{table}[H]
\centering
\caption{Initial learning rate (lr) and batch size used to train the models on the synthetic dataset}
\label{table:synthetic_parameters}
\begin{tabular}{@{}lccl@{}}
\toprule
           & \multicolumn{1}{l}{Bi-LSTM} & \multicolumn{1}{l}{CNN} & Transformer \\ \midrule
lr         & $4\text{e-}3$                      & $1\text{e-}3$                    & $1\text{e-}3$        \\
batch size & 128                         & 128                     & 128         \\ \bottomrule
\end{tabular}
\end{table}

\footnotetext[1]{The encoder layer is the one defined as part of the pytorch library: \url{https://pytorch.org/docs/stable/generated/torch.nn.TransformerEncoderLayer.html}}

\subsection{Models trained on the the ECG dataset}
The configurations of the models trained on the ECG dataset are presented in tables \ref{table:bilstm_parameters_ecg}, \ref{table:cnn_parameters_ecg}, \ref{table:transformer_parameters_ecg}  while the the batch size and learning rate used are presented in table~\ref{table:ecg_parameters}.
\begin{table}[H]
\centering
\caption{Bi-LSTM Models' hyperparameters for the ECG datasets}
\label{table:bilstm_parameters_ecg}
\begin{tabular}{@{}lll@{}}
\toprule
\# & Layer   & Parameters                             \\ \midrule
1  & Bi-LSTM & Units = 64   \\
2  & Dense   & Units =64, Activation = ReLU   \\
3  & BatchNorm   &    \\
4  & Bi-LSTM & Units = 128   \\
5  & Dense   & Units =128, Activation = ReLU   \\
6  & BatchNorm   &    \\
7  & Bi-LSTM & Units = 128   \\
8  & Dense   & Units =128, Activation = ReLU   \\
9  & BatchNorm   &    \\
10  & Bi-LSTM & Units = 128   \\
11 & Dense   & Units =128, Activation = ReLU   \\
12  & BatchNorm   &    \\
13  & Dense   & Units =256, Activation = ReLU   \\
14  & Dense   & Units =2    \\ \bottomrule
\end{tabular}
\end{table}

\begin{table}[H]
\centering
\caption{CNN Models' hyperparameters for the ECG dataset}
\label{table:cnn_parameters_ecg}
\begin{tabular}{@{}lll@{}}
\toprule
\# & Layer   & Parameters                             \\ \midrule
1  & Conv 1D & $128 \times 11$, strides =1   \\                           
2  & Conv 1D & $128 \times 11$, strides =1 \\                        
3  & Conv 1D & $128 \times 11$, strides =1 \\      
4  & Average Pooling &                            \\
5  & Dense   & units =2             \\ \bottomrule
\end{tabular}
\end{table}

\begin{table}[H]
\centering
\caption{Transformer Models' hyperparameters for the ECG datasets}
\label{table:transformer_parameters_ecg}
\begin{tabular}{@{}cll@{}}
\toprule
\# & Layer   & Parameters                             \\ \midrule
1  & Conv 1D & $128 \times 3$, strides =1   \\
1  & Conv 1D & $64 \times 3$, strides =1   \\
3  & Positional encoding &                             \\
4  & Concatenate & Concatenate input and positional embedding  \\
5  & Transformer  Layer\footnotemark[1] &  heads =4, dim. feedforward =256, dropout =0.1, mlp dim = 128  \\
6  & Transformer  Layer\footnotemark[1] & heads  =4, dim. feedforward =256, dropout =0.1, mlp dim = 128  \\
7  & Transformer  Layer\footnotemark[1]& heads =4, dim. feedforward =256, dropout =0.1  , mlp dim = 128  \\
8  & Transformer  Layer\footnotemark[1] &heads =4, dim. feedforward =256, dropout =0.1  , mlp dim = 128  \\  
9  & Dense   & Units =128, Activation = ReLU   \\
10  & Dense   & Units =2    \\ \bottomrule
\end{tabular}
\end{table}

\begin{table}[H]
\centering
\caption{Initial learning rate (lr) and batch size used to train the models on the ecg dataset}
\label{table:ecg_parameters}
\begin{tabular}{@{}lccl@{}}
\toprule
           & \multicolumn{1}{l}{Bi-LSTM} & \multicolumn{1}{l}{CNN} & Transformer \\ \midrule
lr         & $1\text{e-}5$                      & 0.001                   & 0.001       \\
batch size & 64                         & 128                     & 128         \\ \bottomrule
\end{tabular}
\end{table}

\footnotetext[1]{The encoder layer is the one defined as part of the pytorch library: \url{https://pytorch.org/docs/stable/generated/torch.nn.TransformerEncoderLayer.html}}

\subsection{Models trained on the the Ford A dataset}
The configurations of the models trained on the Ford A dataset are presented in tables \ref{table:bilstm_parameters_forda}, \ref{table:cnn_parameters_forda}, \ref{table:transformer_parameters_forda}  while the the batch size and learning rate used are presented in table~\ref{table:forda_parameters}.
\begin{table}[H]
\centering
\caption{Bi-LSTM Models' hyperparameters for the Ford A datasets}
\label{table:bilstm_parameters_forda}
\begin{tabular}{@{}lll@{}}
\toprule
\# & Layer   & Parameters                             \\ \midrule
1  & Bi-LSTM & Units = 64   \\
2  & Dense   & Units =64, Activation = ReLU   \\
3  & BatchNorm   &    \\
4  & Bi-LSTM & Units = 64   \\
5  & Dense   & Units =64, Activation = ReLU   \\
6  & BatchNorm   &    \\
7  & Bi-LSTM & Units = 64   \\
8  & Dense   & Units =64, Activation = ReLU   \\
9  & BatchNorm   &    \\
10  & Bi-LSTM & Units = 64   \\
11 & Dense   & Units =64, Activation = ReLU   \\
12  & BatchNorm   &    \\
13  & Dense   & Units =128, Activation = ReLU   \\
14  & Dense   & Units =2    \\ \bottomrule
\end{tabular}
\end{table}

\begin{table}[H]
\centering
\caption{CNN Models' hyperparameters for the Ford A dataset}
\label{table:cnn_parameters_forda}
\begin{tabular}{@{}lll@{}}
\toprule
\# & Layer   & Parameters                             \\ \midrule
1  & Conv 1D & $128 \times 15$, strides =1   \\                          
2  & Conv 1D & $128 \times 15$, strides =1 \\                         
3  & Conv 1D & $32 \times 15$, strides =1 \\                           
4  & Conv 1D & $128 \times 15$, strides =1 \\
5  & Average Pooling &                            \\
6  & Dense   & units =2             \\ \bottomrule
\end{tabular}
\end{table}

\begin{table}[H]
\centering
\caption{Transformer Models' hyperparameters for the Ford A datasets}
\label{table:transformer_parameters_forda}
\begin{tabular}{@{}cll@{}}
\toprule
\# & Layer   & Parameters                             \\ \midrule
1  & Conv 1D & $128 \times 3$, strides =1   \\
1  & Conv 1D & $64 \times 3$, strides =1   \\
3  & Positional encoding &                             \\
4  & Concatenate & Concatenate input and positional embedding  \\
5  & Transformer  Layer\footnotemark[1] &  heads =4, dim. feedforward =128, dropout =0.026, mlp dim = 256  \\
6  & Transformer  Layer\footnotemark[1] & heads  =4, dim. feedforward =128, dropout =0.026, mlp dim = 256  \\
7  & Transformer  Layer\footnotemark[1]& heads =4, dim. feedforward =128, dropout =0.026  , mlp dim = 256  \\
8  & Transformer  Layer\footnotemark[1] &heads =4, dim. feedforward =128, dropout =0.026  , mlp dim = 256  \\  
9  & Dense   & Units =128, Activation = ReLU   \\
10  & Dense   & Units =2    \\ \bottomrule
\end{tabular}
\end{table}

\begin{table}[H]
\centering
\caption{Initial learning rate (lr) and batch size used to train the models on the Ford A dataset}
\label{table:forda_parameters}
\begin{tabular}{@{}lccl@{}}
\toprule
           & \multicolumn{1}{l}{Bi-LSTM} & \multicolumn{1}{l}{CNN} & Transformer \\ \midrule
lr         & $1.80\text{e-}3$            & $1.61\text{e-}3$      & $1.06\text{e-}5$       \\
batch size & 64                         & 64                     & 64         \\ \bottomrule
\end{tabular}
\end{table}

\clearpage
\section{Classification Metrics}
Each  dataset was split with a 0.7,0.15,0.15 split between the train, validation and test set.  Accuracy along the precision and recall are presented for the classification task on the three datasets 
\label{sec:classification_metrics}
\subsection{Synthetic dataset}

\begin{table}[htb]

\centering
\caption{Classification metrics for the 3 models trained on the synthetic dataset reported for the train, validation (valid.) and test set}
\label{table:metrics_synthetic}
\begin{tabular}{@{}llll|lll|lll@{}}
\toprule
          & \multicolumn{3}{c}{Bi-LSTM}    & \multicolumn{3}{c}{CNN}   & \multicolumn{3}{c}{Transformer} \\ \midrule
          & Train & Valid. & Test  & Train & Valid. & Test  & Train   & Valid.   & Test   \\
Accuracy  & 0.994 & 0.944 & 0.942    & 0.977 & 0.955 & 0.958      & 0.996   & 0.944     & 0.929  \\
Precision & 0.994 & 0.944 & 0.945    & 0.978 & 0.953 & 0.957      & 1       & 0.985     & 0.978  \\
Recall    & 0.993 & 0.938 & 0.935    & 0.974 & 0.952 & 0.957      & 0.991   & 0.895     & 0.875  \\
F1        & 0.994 & 0.941 & 0.940    & 0.976 & 0.953 & 0.957      & 0.995   & 0.938     & 0.923  \\ \bottomrule
\end{tabular}

\end{table}

\subsection{ECG datasets}
\begin{table}[htb]
\centering
\caption{Classification metrics for the 3 models trained on the ECG dataset reported for the train, validation (valid.) and test set}
\label{table:metrics_ECG}
\begin{tabular}{@{}llll|lll|lll@{}}
\toprule
          & \multicolumn{3}{c}{Bi-LSTM}    & \multicolumn{3}{c}{CNN}   & \multicolumn{3}{c}{Transformer} \\ \midrule
          & Train & Valid. & Test  & Train & Valid. & Test  & Train   & Valid.   & Test   \\
Accuracy  & 0.941 & 0.942 & 0.939    & 0.960 & 0.947 & 0.958      & 0.951   & 0.948        & 0.957  \\
Precision & 0.882 & 0.880 & 0.866    & 0.939 & 0.904 & 0.936      & 0.915   & 0.902        & 0.922  \\
Recall    & 0.908 & 0.905 & 0.896    & 0.915 & 0.895 & 0.896      & 0.906   & 0.902        & 0.908  \\
F1        & 0.895 & 0.892 & 0.881    & 0.927 & 0.899 & 0.916      & 0.910   & 0.902        & 0.915  \\ \bottomrule
\end{tabular}
\end{table}
\newpage
\subsection{Ford A datasets}
\begin{table}[htb]
\centering
\caption{Classification metrics for the 3 models trained on the Ford A dataset reported for the train, validation (valid.) and test set}
\label{table:metrics_FordA}
\begin{tabular}{@{}llll|lll|lll@{}}
\toprule
          & \multicolumn{3}{c}{Bi-LSTM}    & \multicolumn{3}{c}{CNN}   & \multicolumn{3}{c}{Transformer} \\ \midrule
          & Train & Valid. & Test  & Train & Valid. & Test  & Train   & Valid.   & Test   \\
Accuracy  & 0.808 & 0.778 & 0.770    & 0.946 & 0.933 & 0.936      & 1   & 0.869 & 0.854  \\
Precision & 0.837 & 0.784 & 0.782    & 0.945 & 0.922 & 0.927      & 1   & 0.860 & 0.860  \\
Recall    & 0.757 & 0.721 & 0.726    & 0.945 & 0.936 & 0.941      & 1   & 0.857 & 0.884  \\
F1        & 0.795 & 0.751 & 0.753    & 0.945 & 0.929 & 0.934      & 1   & 0.858 & 0.872  \\ \bottomrule
\end{tabular}
\end{table}

\section{Theoretical estimate for $\tilde{S}_A-$TIC curves}
\label{sec:theoretical_estimates}
Given an attribution scheme $\mathcal{A}$ which assigns relevance $\mathbf{R}$ to the input $\mathbf{X}$ for a specific class $c \in C$ such that $\mathcal{A}_{c}: \mathbf{X} \rightarrow \left \{\mathbf{R} \in \mathbb{R}^{M \times T } \right \}$, where $\mathbf{X} = \left(x_{m,t} \right)$ and  $\mathbf{R} = \left(r_{m,t} \right)$, with $m$ and $t$ being the indices associated to the number of features $M$ and to the number of time steps $T$, respectively. The following subsets of the relevance can be defined as:
\begin{subequations}
\begin{align}
    R^+ &= \left \{ r_{m,t} \vert r_{m,t}>0   \right \} \\
    R^- &= \left \{ r_{m,t} \vert r_{m,t}<0   \right \} \\
    R^+_k &= \left \{ r_{m,t} \vert r_{m,t} \in R^+ \cap  r_{m,t} \geq Q_R(1-k)  \right \} \\
    R^{+C}_k &= \left \{ r_{m,t} \vert r_{m,t} \in R^+ \cap  r_{m,t} < Q_R(1-k)  \right \} \label{eq.sum_pos}
\end{align}
\end{subequations}
The previous subsets can then be combined as follows:
\begin{subequations}
\begin{align}
    \sum_R r_{m,t} &= \sum_{R^+} r_{m,t} + \sum_{R^-} r_{m,t} \label{subeq:R_1} \\
    \sum_{R^+} r_{m,t} &= \sum_{R^+_k} r_{m,t} +\sum_{R^{+C}_k} r_{m,t} \label{subeq:R_2}
\end{align}
\end{subequations}
where eq.~\ref{subeq:R_2} is the sum of the relevance in the set $R^+_k$ which represents the relevance assigned to corrupted time steps in sample $\mathbf{\bar{X}}^{\text{top}}_k$ and the complement set $R^{+C}_k$ represents the sum of the positive relevance assigned to uncorrupted time steps.
The theoretical estimate for the $\tilde{S}_A-$TIC curves can be estimated starting from the initial definitions for $\tilde{S}_A$ and TIC: 
\begin{align}
    \tilde{S}_A(k) &=   \frac{S\left(\mathbf{X}\right)- S\left(\mathbf{\bar{X}}^{\text{top}}_k\right)}{S\left(\mathbf{X}\right)- S\left(\mathbf{\bar{X}}^{\text{top}}_{k=1}\right)} \label{eq.sup_SA} \\
    \text{TIC}(k) &= \frac{\sum_{R^+_k} r_{m,t}}{\sum_{R^+} r_{m,t} + \epsilon}  \label{eq.sup_TIC}
\end{align}
For the rest of the analysis the constant $\epsilon$ is discarded as this constant is only used to prevent the equation to go to infinity.
Using the linear additivity property we can relate the models' output to the relevance as follows: \begin{subequations}
    \begin{align}
        S\left(\mathbf{X}\right) &= \sum_{R^+} r_{m,t} + \sum_{R^-} r_{m,t} + \phi_0 \label{subeq:S_1} \\
        S\left(\mathbf{\bar{X}}^{\text{top}}_{k=1}\right) &=  \sum_{R^-} r_{m,t} + \phi_o \label{subeq:S_2} \\
        S\left(\mathbf{\bar{X}}^{\text{top}}_{k}\right) &= \sum_{R^-} r_{m,t} + \sum_{R^{+C}_k} r_{m,t} + \phi_o \label{subeq:S_3}
    \end{align}
\end{subequations}
where $\phi_0$ is a constant, which is equal to the expectancy of the network for Shapley-based methods. 
Combining eq.~\ref{subeq:S_1} and eq.~\ref{subeq:S_2}: \begin{equation}
    S(\mathbf{X})- S(\mathbf{\bar{X}}^{\text{top}}_{k=1}) = \sum_{R^+} r_{m,t} \label{eq.bottom_frac}
\end{equation}
and combining eq.~\ref{subeq:S_1} and eq.~\ref{subeq:S_3}: \begin{equation}
    S\left(\mathbf{X}\right)- S\left(\mathbf{\bar{X}}^{\text{top}}_{k}\right) = \sum_{R^+} r_{m,t} - \sum_{R^{+C}_k} r_{m,t} = \sum_{R^+_k} r_{m,t} \label{eq.top_frac}
\end{equation} 
Substituting eq.~\ref{eq.bottom_frac} and~\ref{eq.top_frac} into eq.~\ref{eq.sup_SA}, we obtain: \begin{equation}
     \tilde{S}_A(k) = \frac{\sum_{R^+_k} r_{m,t}}{\sum_{R^+} r_{m,t} + \epsilon} = \text{TIC}
\end{equation}
showing that  $\tilde{S}_A(k)$, which measures the change in score as a sample $\mathbf{X}$ is corrupted, is related to the fraction of positive relevance corrupted measured with the $\text{TIC}$ index.

\end{document}